\documentclass[10pt]{article} 

\usepackage[preprint]{rlj} 

\usepackage{amssymb}            
\usepackage{mathtools}          
\usepackage{mathrsfs}           
\usepackage{graphicx}           
\usepackage{subcaption}         
\usepackage[space]{grffile}     
\usepackage{url}                
\usepackage{lipsum}             

\title{Reward-Zero: Language Embedding Driven Implicit Reward Mechanisms for Reinforcement Learning}

\setrunningtitle{Reward-Zero: Language Embedding Implicit Reward Mechanisms for Reinforcement Learning}

\author{Heng Zhang\textsuperscript{1,2,3}, Haddy Alchaer\textsuperscript{4}, Arash Ajoudani\textsuperscript{1}, Yu She\textsuperscript{3,4,5}}

\emails{heng.zhang@iit.it}

\affiliations{
$^{1}$\textbf{Human-Robot Interfaces and Interaction Lab, Istituto Italiano di Tecnologia, Genova, Italy}\\
$^{2}$\textbf{Ph.D. program of national interest in Robotics and Intelligent Machines (DRIM) and Universit `a di Genova, Genoa, Italy.}\\
$^{3}$\textbf{Edwardson School of Industrial Engineering, Purdue University, West Lafayette, IN, USA.}\\
$^{4}$\textbf{Elmore Family School of Electrical and Computer Engineering, Purdue University, West Lafayette, IN, USA.}\\
$^{5}$\textbf{School of Mechanical Engineering, Purdue University, West Lafayette, IN, USA.}}

\contribution{
    We propose Reward-Zero, an implicit reward mechanism that uses CLIP vision-language embeddings with a baseline-penalty potential to produce dense completion-sense signals from a natural-language goal description and raw visual observations, without task-specific reward engineering.
    }
    {
    Prior language-guided reward methods rely on VLM captioning ($\sim$2\,s/frame) or LLM reward-code synthesis; Reward-Zero operates via direct embedding comparison ($\sim$5\,ms/frame).
    }

\contribution{
    We introduce a completion-sense mini benchmark that evaluates whether language-grounded reward models assign monotonically increasing potentials across task-completion stages, and use it to show that CLIP-direct with baseline penalty (72\% forward accuracy, perfect jump detection) outperforms VLM-caption pipelines (67\% best) while being 400$\times$ faster.
    }
    {
    This benchmark isolates reward-signal fidelity from RL optimization dynamics, which existing evaluations typically conflate.
    }

\contribution{
    We show that Reward-Zero, integrated as an auxiliary reward into PPO, can accelerate convergence and improve success rates on ManiSkill robotic tasks compared to the PPO baseline with hand-crafted dense rewards alone.
    }
    {
    Same PPO hyperparameters and environment configurations as baselines; the only difference is the addition of the Reward-Zero auxiliary signal.
    }


\summary{
Reinforcement learning agents often struggle with sparse or poorly shaped reward signals, and hand-crafting dense rewards for each new task is labor-intensive and error-prone. We introduce Reward-Zero, an implicit reward mechanism that derives dense progress signals from natural-language goal descriptions using pre-trained vision-language embeddings. Given only a textual goal (e.g., ``The cabinet drawer is fully open'') and raw visual observations, Reward-Zero computes a potential function via CLIP image--text similarity with a baseline-penalty term that penalizes visual similarity to the initial state. The resulting reward is continuous, deterministic, and computable in $\sim$5\,ms per frame, enabling dense per-step feedback during online RL training without task-specific engineering.

We validate Reward-Zero in two stages. First, we develop a completion-sense mini benchmark on ManiSkill simulation tasks, showing that CLIP-direct with baseline penalty achieves 72\% forward transition accuracy and perfect jump detection (6/6 episodes), outperforming VLM-caption pipelines (67\% best) while being 400$\times$ faster. Second, we integrate Reward-Zero as an auxiliary reward into PPO for robotic manipulation and locomotion tasks, demonstrating faster convergence, more stable training dynamics, and higher success rates compared to the PPO baseline with hand-crafted dense rewards alone.
}

\begin{document}

\makeCover  
\maketitle  

\begin{abstract}
We introduce Reward‑Zero, a general-purpose implicit reward mechanism that transforms natural-language task descriptions into dense, semantically grounded progress signals for reinforcement learning (RL). Reward‑Zero serves as a simple yet sophisticated universal reward function that leverages language embeddings for efficient RL training. By comparing the embedding of a task specification with embeddings derived from an agent’s interaction experience, Reward‑Zero produces a continuous, semantically aligned sense-of-completion signal. This reward supplements sparse or delayed environmental feedback without requiring task-specific engineering. When integrated into standard RL frameworks, it accelerates exploration, stabilizes training, and enhances generalization across diverse tasks. Empirically, agents trained with Reward‑Zero converge faster and achieve higher final success rates than conventional methods such as PPO with common reward-shaping baselines, successfully solving tasks that hand-designed rewards could not in some complex tasks. In addition, we develop a mini benchmark for evaluation of completion sense during task execution via language embeddings. These results highlight the promise of language-driven implicit reward functions as a practical path toward more sample-efficient, generalizable, and scalable RL for embodied agents.
Code will be released after peer review.
\end{abstract}

\section{Introduction }
\label{sec:intro}
\begin{quote}
    \textit{``Die Grenzen meiner Sprache bedeuten die Grenzen meiner Welt."}\\
    \hfill \textbf{------ Wittgenstein, 1922}
\end{quote}
Reinforcement learning (RL) has shown remarkable potential in a broad range of domains, from robotic manipulation~\cite{kalashnikov2018scalable,10517611} and strategic game playing~\cite{silver2016mastering} to autonomous driving~\cite{kendall2019learning}. 
Its core promise lies in enabling agents to learn complex behaviors directly through interaction with their environments~\cite{sutton2018reinforcement}, offering a pathway toward adaptive, intelligent systems applicable to real-world problems. 
However, the success of RL critically depends on the design of effective reward functions that guide the learning process~\cite{yu2025reward}. Engineering these rewards for non-trivial tasks is often a challenging, time-consuming, and error-prone endeavor. 
Hand-crafted reward signals may capture only partial aspects of a desired behavior, leading to unintended incentives or misaligned learning objectives. In this age of increasingly complex tasks and open-ended environments \cite{silver2025welcome}, 
developing more scalable and generalizable reward mechanisms has become a central challenge for advancing modern RL~\cite{goyal2019using}.

Reward design lies at the heart of RL, as it defines the objective that governs agent~\cite{yu2025reward}. Poorly designed rewards can yield brittle or misaligned policies, drastically slowing convergence or encouraging unintended strategies. 
In many real-world tasks, obtaining accurate, detailed feedback is extremely difficult, as environments may provide only sparse or delayed signals, 
and hand-engineered rewards often fail to reflect nuanced notions of progress or partial success. 
Consequently, researchers have sought more generalizable paradigms for deriving rewards that are grounded in semantic understanding rather than hand-tuned~\cite{language_reward_models2025}. 

Traditional approaches to improving reward signals include techniques such as reward shaping, curriculum learning, and auxiliary objectives. Reward shaping methods, for instance, transform sparse environmental rewards into denser signals by adding heuristic-based intermediate rewards~\cite{wiewiora2003potential,11128552}. Although this can accelerate learning, such methods often require extensive domain knowledge and manual tuning. Potential-based shaping provides theoretical guarantees of policy invariance but offers limited flexibility in capturing complex semantic relationships between actions and goals~\cite{wiewiora2003potential,trainable_pbrs2026}. These challenges highlight the limitations of purely task-specific designs and motivate the need for a more general, adaptive approach. Ideally, a universal reward mechanism should provide semantically coherent guidance without heavy engineering effort, enabling agents to learn from natural task descriptions and adapt across diverse environments with minimal redesign.

Recent advances in language modeling and representation learning offer a compelling direction~\cite{language_reward_models2025,langreward2023,zhang2025rewind}: natural language encapsulates high-level task semantics and contextual intent, suggesting that linguistic structures could serve as a basis for flexible and informative reward mechanisms applicable across domains.
However, existing language-guided reward methods often rely on VLM captioning or LLM reward-code synthesis, which can be computationally expensive and may suffer from issues such as goal-echo bias or brittle grounding. 
Moreover, these methods typically require additional engineering to ensure that the generated rewards are informative and stable during training.

\begin{figure}[h]
    \centering
\includegraphics[width=1\linewidth]{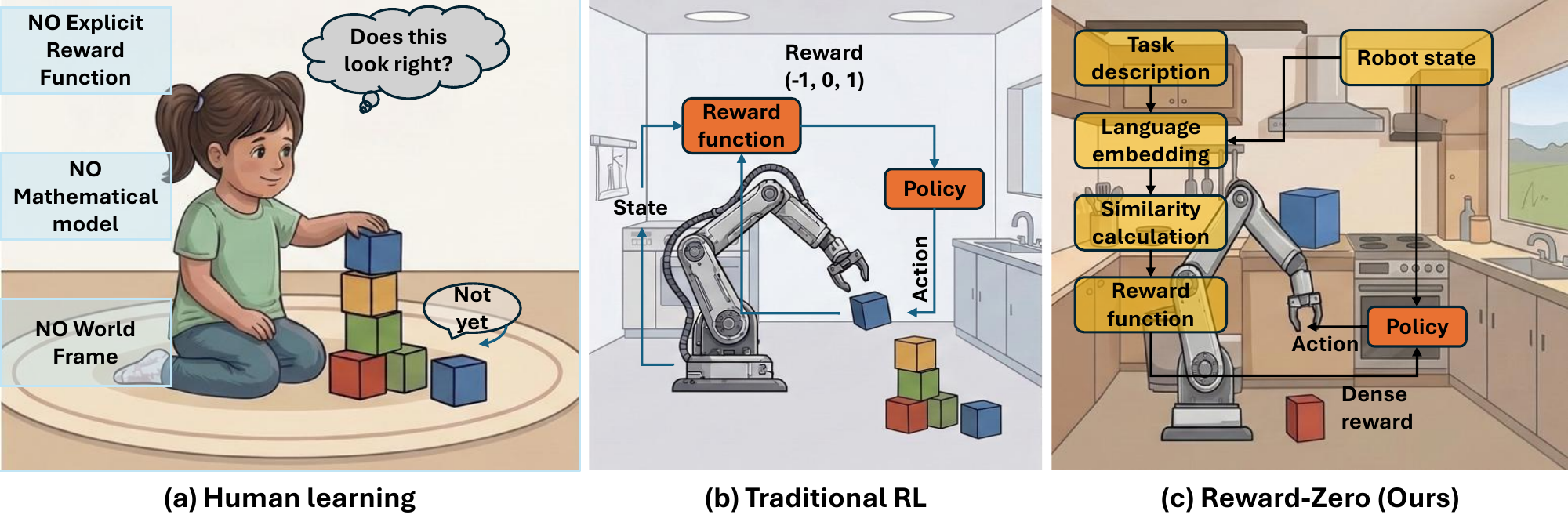}
    \caption{Conceptual comparison of human learning, traditional RL, and the proposed Reward-Zero. 
    The left panel illustrates how Human Learning is intuitive and implicit, driven by visual matching and a generalized "Sense of Completion," without relying on explicit rewards, mathematical environment models, or exact object world coordinates. 
    The center panel depicts Traditional RL training a simple robotic policy, a rigid and explicit approach that typically requires a hand-crafted reward function, a precise environment model, heavy observation and exact object coordinates to function. 
    The right panel introduces Reward-Zero, our flexible and language-driven method that represents a sophisticated universal reward function. 
    This mechanism uses a general-purpose language embedding-driven implicit reward mechanism to generate a continuous sense-of-completion signal by comparing task and experience embeddings. As shown, Reward-Zero aims to eliminate hand-crafted rewards by relying only on raw language embedding, enhancing generalization across diverse tasks.
    \textbf{Zero} here signifies the absence of hand-crafted rewards, without explicit reward engineering. This is the \textbf{Zero} step toward more general, adaptable, and scalable RL that can learn from natural language descriptions and raw observations, much like humans do.
    }
    \label{fig:motivation}
\end{figure}

To address these challenges, we propose \textbf{Reward‑Zero}, see Fig.~\ref{fig:motivation}, a general-purpose implicit reward mechanism that transforms natural-language task descriptions into continuous progress signals.
Inspired by the human ability to intuitively gauge task completion from semantic understanding, Reward‑Zero leverages pre-trained language embeddings to encode both task instructions and the agent’s ongoing experience, computing a semantically aligned "sense of completion" signal that evolves during learning. This implicit reward serves as a universal auxiliary objective that complements or substitutes sparse environmental feedback, removing the need for hand-crafted shaping functions. By grounding reward estimation in language semantics, Reward‑Zero captures nuanced progress even without explicit task-specific signals. The resulting approach provides a scalable route for improving exploration efficiency, training stability, and generalization across varied domains from simulated manipulation tasks to embodied, real-world agents interacting in open environments.

This paper makes three primary contributions. First, we introduce the Reward‑Zero, a language embedding–driven implicit reward mechanism. Second, we propose a mini benchmark designed to quantitatively assess an agent’s "completion sense", how well language-grounded feedback tracks progress during task execution. Third, we present comprehensive empirical evaluations demonstrating that Reward‑Zero outperforms conventional algorithms such as PPO and standard reward-shaping baselines in terms of convergence speed, stability, and final performance. Through these contributions, our work highlights the promise of language-embedding implicit rewards as a practical step toward more generalizable and sample-efficient RL.

\section{Related Work}
\subsection{Potential-Based Reward Function}
Recent advances in potential-based reward shaping (PBRS) have extended the original theoretical guarantees to richer, trainable shaping functions and intrinsic‑motivation signals, improving sample efficiency in sparse‑reward tasks while preserving optimal‑policy invariance under broader conditions \cite{ng1999policy,wiewiora2003potential,mueller2025improving,trainable_pbrs2026}. Work integrating PBRS with intrinsic curiosity and count‑based exploration has enabled faster exploration in procedurally complex environments \cite{pathak2017curiosity,bellemare2016unifying,xu2025stare}. At the same time, practical limitations remain: many modern intrinsic rewards are nonstationary or depend on learned representations, which can violate PBRS assumptions and induce policy bias; computational overhead and sensitivity to potential design also hinder transferability across domains \cite{sutton2018reinforcement,schmidhuber1991curiosity,houthooft2016vime,mueller2025improving}. Recent methods therefore focus on converting learned intrinsic signals into potential‑based forms or constraining their dependence to preserve optimality, but trade‑offs between theoretical guarantees and empirical performance persist \cite{11128552,trainable_pbrs2026}.

\subsection{Language-Guided Reward for RL}
Language-guided reward learning uses natural-language specifications and large language models to produce dense, interpretable reward signals that accelerate learning in sparse, compositional, or multi-step tasks. Recent work demonstrates automatic synthesis of shaped rewards from textual goals, enabling data‑efficient policy learning and iterative human refinement for robotics and embodied benchmarks \cite{text2reward2023,zhang2025rewind}. Integrations that fine‑tune or prompt LLMs to generate reward functions or reward classifiers show improved generalization across task variants and reduce the need for new demonstrations \cite{lmgt2024,language_reward_models2025}. Other lines pretrain joint policy–reward models on language‑annotated trajectories to transfer to unseen instructions \cite{langreward2023}. Remaining challenges include brittle grounding when environment observations misalign with textual abstractions, latent biases in LLM‑derived rewards, and the need for verification to prevent reward hacking; recent proposals therefore emphasize human‑in‑the‑loop refinement, robust grounding mechanisms, and formalizing safety guarantees for language‑derived rewards \cite{zhang2025rewind,llm_reward_align2026,language_reward_models2025}.


\section{Method}

In this section, we present Reward-Zero, a vision-language model-based reward computation approach that provides dense, semantically grounded rewards for robotic manipulation tasks. Our method consists of three key components: (1) language embedding-based potential estimation, (2) progress-aware activation, and (3) completion-sense reward shaping.

\subsection{Language Embedding-Based Potential Estimation}

Traditional reward shaping in robotics often relies on hand-crafted distance metrics or task-specific state features, which limits generalization across diverse manipulation tasks. We propose a fundamentally different approach that leverages the semantic understanding capabilities of vision-language models (VLMs) to compute task-relevant potentials directly from natural language descriptions.

Our core insight is that the semantic similarity between a scene description and a goal description naturally captures task progress without requiring explicit geometric or kinematic computations. Specifically, we define the potential function $\Phi(s)$ as the cosine similarity between the encoded current state caption and the encoded goal description:

\begin{equation}
\Phi(s) = \cos(\mathbf{e}_{\text{state}}, \mathbf{e}_{\text{goal}}) = \frac{\mathbf{e}_{\text{state}} \cdot \mathbf{e}_{\text{goal}}}{\|\mathbf{e}_{\text{state}}\| \|\mathbf{e}_{\text{goal}}\|}
\end{equation}

where $\mathbf{e}_{\text{state}}$ and $\mathbf{e}_{\text{goal}}$ are the text embeddings of the current scene caption and goal description, respectively. This formulation yields a bounded potential $\Phi(s) \in [-1, 1]$, where higher values indicate greater semantic alignment with the goal.

To enhance the expressiveness and discriminability of our potential function, we employ an enrichment procedure for both state captions and goal descriptions. Rather than using raw, terse captions (e.g., ``robot arm near cup''), we leverage large language models to generate detailed, contextually-rich descriptions that capture nuanced aspects of the scene. For state descriptions, the VLM captioner is prompted to produce comprehensive scene descriptions that include object positions, spatial relationships, gripper states, and ongoing actions. Similarly, goal descriptions are enriched to include expected final configurations, success criteria, and relevant contextual details.

This enrichment process is crucial for two reasons. First, richer text provides more distinctive embeddings in the semantic space, enabling finer-grained progress discrimination. Second, detailed descriptions help bridge the semantic gap between visual observations and abstract task specifications, allowing the embedding space to capture task-relevant features more effectively. The generality of this approach stems from its reliance solely on pretrained language models and text encoders, requiring no task-specific engineering or domain knowledge.

\subsection{Progress-Aware Activation}

While the potential function provides a continuous measure of goal proximity, raw potential differences may not adequately incentivize the agent during critical final stages of task completion. We introduce a progress-aware activation mechanism that dynamically amplifies rewards as the agent approaches task completion.

The activation function is based on a sigmoid transformation centered at a completion threshold $\tau$:

\begin{equation}
\sigma_{\text{act}}(\Phi) = \frac{1}{1 + \exp(-k \cdot (\Phi - \tau))}
\end{equation}

where $k$ controls the steepness of the activation transition and $\tau$ represents the potential value at which the agent is considered to be near completion. This sigmoid activation provides several desirable properties: (1) it remains near zero when the agent is far from the goal, avoiding interference with exploration; (2) it smoothly transitions to high values as the agent approaches completion, providing increasingly strong guidance; and (3) it avoids discontinuous reward jumps that could destabilize learning.

To further encourage sustained progress during the critical completion phase, we incorporate a progress multiplier that rewards continued improvement:

\begin{equation}
\Delta\Phi = \max(0, \Phi_t - \Phi_{t-1})
\end{equation}

The progress term ensures that the agent receives additional reward for making forward progress even when already close to the goal. This addresses a common challenge in potential-based shaping where the reward signal diminishes as the agent approaches the goal state, potentially leading to sluggish final movements. By multiplying the activation by $(1 + \Delta\Phi)$, we create a reward landscape that actively pulls the agent toward completion rather than merely indicating proximity.

\subsection{Completion-Sense Reward Formulation}

Combining the language-based potential with progress-aware activation, we formulate the complete Reward-Zero function as:

\begin{equation}
\label{eq:completion-reward}
R_{\text{completion}} = r_{\text{base}} + \beta \cdot \sigma_{\text{act}}(\Phi) \cdot (1 + \Delta\Phi)
\end{equation}

where $r_{\text{base}}$ represents the base potential reward (which can be $\Phi$ itself or the potential difference $\Phi_t - \Phi_{t-1}$), $\beta$ is the completion bonus weight, $\sigma_{\text{act}}(\Phi)$ is the sigmoid activation function, and $\Delta\Phi$ captures the instantaneous progress.

This formulation achieves several design objectives that make it effective for general robotic manipulation:

\textbf{Smoothness and Stability:} Unlike sparse rewards or threshold-based completion bonuses, our reward function is continuous and differentiable everywhere. This property is essential for stable policy gradient estimation and avoids the optimization difficulties associated with discontinuous reward landscapes.

\textbf{Automatic Curriculum:} The reward naturally implements a form of curriculum learning. Early in training, when the agent is far from the goal, the completion bonus remains inactive, allowing the base potential to guide exploration. As the agent learns to approach the goal, the completion bonus gradually activates, providing stronger incentives for precise final positioning.

\textbf{Task Generalization:} By grounding rewards in language semantics rather than task-specific metrics, our approach generalizes across diverse manipulation tasks without modification. The same reward function can evaluate progress toward ``pick up the red cube,'' ``close the drawer,'' or ``stack blocks in order'' simply by changing the goal text description.

\textbf{Dense Feedback:} Every timestep receives meaningful reward signal derived from semantic scene understanding, addressing the sparse reward challenge that plagues many manipulation tasks. This density accelerates learning and provides consistent gradient information throughout training.

The hyperparameters $\tau$, $k$, and $\beta$ can be tuned based on task characteristics, though we find that default values of $\tau=0.7$, $k=10$, and $\beta=0.5$ work well across a variety of manipulation scenarios. The threshold $\tau$ should be set below the expected completion potential to ensure the bonus activates during the approach phase rather than only at completion.

\section{Experiments and Results}
\label{sec:exp}
In this section, we evaluate the effectiveness of our proposed Reward-Zero across a range of robotic manipulation tasks. 
We compare our method against standard RL algorithms and common reward-shaping baselines to demonstrate its advantages in terms of sample efficiency, convergence speed, and final performance (success rate).
First, we develop a mini benchmark for evaluation of completion sense during task execution via language embeddings, see Section~\ref{sec:bench}.
Then, we evaluate our method on a suite of robotic manipulation tasks that require varying levels of complexity and dynamic interaction, see Section~\ref{sec:exp-robot}.

\subsection{Mini Benchmark for Completion-Sense Evaluation} \label{sec:bench}

Before deploying Reward-Zero as an auxiliary reward signal in online RL training, we must verify that the underlying potential function $\Phi(s)$ reliably tracks task-completion progress from visual observations alone. To this end, we develop a lightweight offline benchmark that evaluates \emph{completion-sense discrimination}: the ability of a reward model to assign monotonically increasing potential values to frames sampled at increasing stages of task completion, given only a natural-language goal description.
\begin{figure}
    \centering
    \includegraphics[width=1\linewidth]{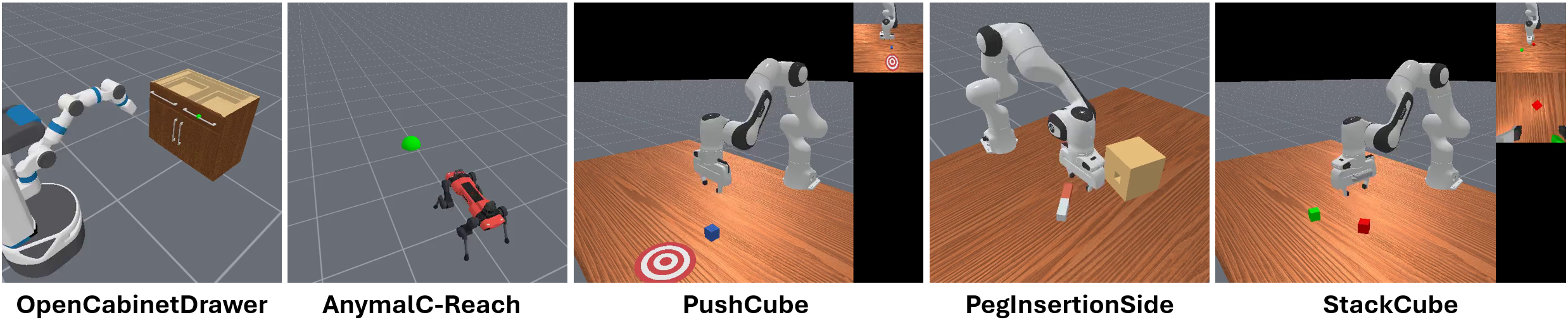}
    \caption{Example tasks and keyframes from the completion-sense mini benchmark. Each episode contains 2--4 annotated keyframes at known completion percentages (0\%, 33\%, 50\%, 66\%, 100\%) extracted from successful ManiSkill~\cite{gu2023maniskill2} rollouts. The benchmark includes tasks with varying visual complexity, from large state changes (e.g., OpenCabinetDrawer) to fine-grained manipulations (e.g., PegInsertionSide).}
    \label{fig:mini-benchmark-tasks}
\end{figure}

\paragraph{Benchmark Design}
We construct evaluation episodes from ManiSkill~\cite{gu2023maniskill2} simulation trajectories across five robotic tasks spanning locomotion, tabletop manipulation, and articulated-object interaction (Fig.~\ref{fig:mini-benchmark-tasks} and Tab.~\ref{tab:bench-tasks}). For each task, we extract keyframes at four known completion percentages (0\%, 33\%, 66\%, 100\%) from successful rollouts, yielding 4 annotated frames per episode. Each task is paired with an end-state goal description (e.g., ``The cabinet drawer is fully open'') rather than an action command, as we found that action-phrased goals (e.g., ``Open the drawer by pulling the handle'') cause VLM-based models to echo goal language even at 0\% completion, inflating similarity scores. In total, the benchmark contains 6 evaluation episodes across 5 environments with 24 annotated keyframes, providing 18 consecutive forward transitions for assessment.

\begin{table}[h]
\centering
\small
\caption{Tasks in the completion-sense mini benchmark. Each episode contains keyframes at known completion percentages extracted from successful ManiSkill rollouts. Two episodes are included for OpenCabinetDrawer to test consistency across different initial configurations.}
\label{tab:bench-tasks}
\begin{tabular}{lccl}
\hline
\textbf{Task} & \textbf{Frames} & \textbf{Completions (\%)} & \textbf{Goal Description} \\
\hline
OpenCabinetDrawer ($\times$2) & 4 & 0, 33, 66, 100 & The cabinet drawer is fully open \\
AnymalC-Reach & 4 & 0, 33, 66, 100 & The quadruped robot is at the target position \\
PushCube & 4 & 0, 33, 66, 100 & The cube is at the target position \\
PegInsertionSide & 4 & 0, 33, 66, 100 & The peg is fully inserted into the hole \\
StackCube & 4 & 0, 33, 66, 100 & The red cube is stacked on the green cube \\
\hline
\end{tabular}
\end{table}

These tasks were selected to span a range of visual complexities: OpenCabinetDrawer and AnymalC-Reach involve large, visually distinctive state changes, while PegInsertionSide and StackCube involve fine-grained manipulations where objects are small relative to the scene.

\paragraph{Evaluation Metrics.}
We evaluate reward models along four complementary dimensions:
\begin{itemize}[leftmargin=1.5em]
    \item \textbf{Forward Transition Accuracy (FTA)}: For each consecutive pair of frames $(s_t, s_{t+1})$ where completion increases, we check whether the computed reward $R(s_t, s_{t+1}) > \epsilon$ (with threshold $\epsilon = 0.001$). This directly measures whether the reward signal provides a positive learning gradient in the direction of task progress. Reported out of 18 total transitions.
    \item \textbf{Monotonicity}: The fraction of consecutive potential pairs $(\Phi(s_t), \Phi(s_{t+1}))$ that strictly increase. A score of 1.0 indicates perfectly monotonic potential tracking across the episode.
    \item \textbf{Spearman Correlation ($\rho$)}: Rank correlation between ground-truth completion percentages and computed potential values, capturing ordinal alignment without assuming linearity.
    \item \textbf{Jump Detection (J+)}: Whether a single 0\%$\rightarrow$100\% transition produces a clearly positive reward ($R > \epsilon$), testing the model's sensitivity to large state changes. Reported out of 6 episodes.
\end{itemize}
Forward transition accuracy and jump detection are the most directly relevant metrics for RL training, where the agent always progresses forward from an initial state.

\paragraph{Methods Compared.}
We evaluate two families of approaches for instantiating the potential function $\Phi(s)$:

\noindent\textit{(1) VLM-caption pipeline.} An image is passed through a vision-language model (VLM) which generates a natural-language scene description. This caption is encoded using a sentence embedding model (MiniLM-L6~\cite{wang2020minilm}), and the cosine similarity between the caption embedding and the goal-text embedding serves as the potential. We test Qwen2.5-VL-3B~\cite{bai2025qwen25vl} with three prompting strategies:
\begin{itemize}[leftmargin=1.5em, topsep=2pt]
    \item \textbf{Progress-description}: The prompt provides the goal text and asks the VLM to describe both the current scene state and progress toward the goal. This is the most informative prompt but risks goal-echo bias.
    \item \textbf{Observe-only}: The prompt asks the VLM to describe the scene without revealing the goal, avoiding echo bias at the cost of less goal-directed captions.
    \item \textbf{Evidence-gating}: The prompt instructs the VLM to enumerate only actions that have been visibly completed, with a fallback phrase (``No actions completed'') if uncertain.
\end{itemize}

\noindent\textit{(2) CLIP-direct.} The image is encoded directly by CLIP's~\cite{radford2021learning} vision encoder (ViT-B/32) and compared against the CLIP text encoding of the goal description. No VLM captioning or intermediate text representation is needed. We define the potential as:
\begin{equation}
\Phi(s) = \alpha \cdot \mathrm{sim}\bigl(f_I(s),\; f_T(g)\bigr) \;-\; (1 - \alpha) \cdot \mathrm{sim}\bigl(f_I(s),\; f_I(s_0)\bigr)
\label{eq:clip-potential}
\end{equation}
where $f_I$ and $f_T$ are CLIP's image and text encoders, $g$ is the goal text, and $s_0$ is the initial observation (episode baseline). The first term measures goal proximity; the second term penalizes visual similarity to the initial state, encouraging departure from the start configuration. We use $\alpha = 0.7$, balancing goal affinity with departure from the initial state.

\paragraph{Results.}
Table~\ref{tab:bench-results} presents the main comparison across all methods. CLIP-direct with baseline penalty ($\alpha = 0.7$) achieves the highest forward transition accuracy (13/18, 72\%) and perfect jump detection (6/6), outperforming all VLM-based approaches.

\begin{table}[h]
\centering
\small
\caption{Mini benchmark results comparing VLM-caption and CLIP-direct approaches on 18 forward transitions across 6 episodes. \textbf{FTA}: forward transition accuracy (out of 18). \textbf{J+}: jump detection, 0\%$\rightarrow$100\% (out of 6). \textbf{Mono}: fraction of episodes with fully monotonic potentials (out of 6). Latency is per-frame inference time on a single A30 GPU. Best results in \textbf{bold}. CLIP results are deterministic.}
\label{tab:bench-results}
\begin{tabular}{lcccc}
\hline
\textbf{Method} & \textbf{FTA} ($\uparrow$) & \textbf{J+} ($\uparrow$) & \textbf{Mono} ($\uparrow$) & \textbf{Latency} \\
\hline
\multicolumn{5}{l}{\textit{VLM-caption pipeline (Qwen2.5-VL-3B + MiniLM)}} \\
\quad + progress-description & 12/18 & 5/6 & 2/6 & $\sim$2\,s \\
\quad + observe-only & 12/18 & \textbf{6/6} & 1/6 & $\sim$2\,s \\
\quad + evidence-gating & 0/18 & 0/6 & 0/6 & $\sim$2\,s \\
\hline
\multicolumn{5}{l}{\textit{CLIP-direct (ViT-B/32), $\alpha=0.7$}} \\
\quad with baseline penalty & \textbf{13/18} & \textbf{6/6} & \textbf{2/6} & $\sim$5\,ms \\
\hline
\end{tabular}
\end{table}

We showcase the per-task keyframes and discussion in Appendix~\ref{app:bench-frames}.

\begin{figure}[h]
\centering
\includegraphics[width=\linewidth]{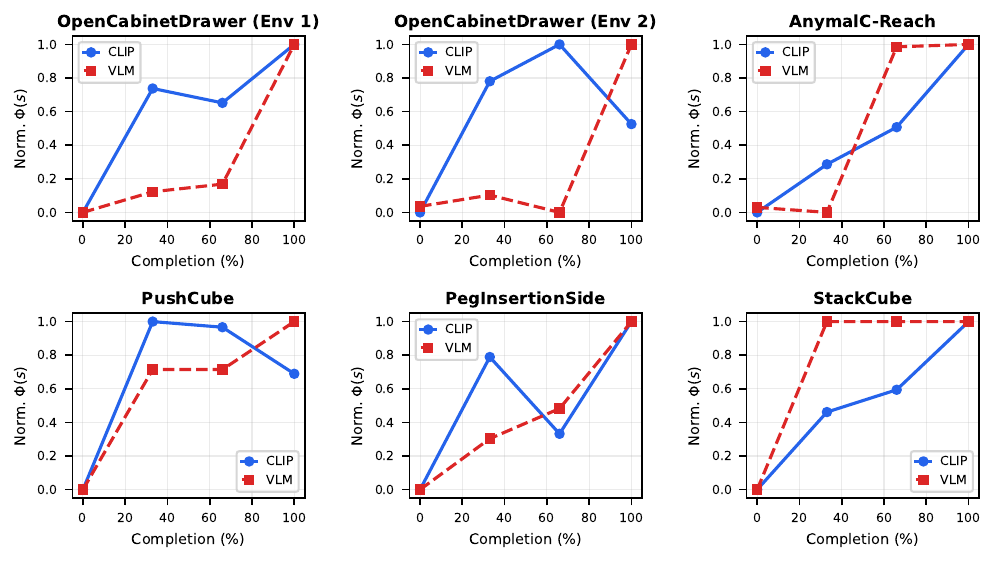}
\caption{CLIP potential $\Phi(s)$ vs.\ task completion (\%) for each benchmark task (CLIP-direct, $\alpha=0.7$). OpenCabinetDrawer shows two episodes (solid/dashed) to illustrate consistency across initial configurations. All tasks now use four keyframes at 0\%, 33\%, 66\%, and 100\% completion. Tasks with large visual changes show strong monotonic trends; fine-manipulation tasks exhibit smaller potential ranges.}
\label{fig:bench-curves}
\end{figure}

\paragraph{Key Findings.}
Several findings emerge from the aggregate comparison (Table~\ref{tab:bench-results}) and per-task analysis (Figures~\ref{fig:bench-frames}--\ref{fig:bench-curves}):

\textbf{CLIP-direct outperforms VLM-caption pipelines on forward discrimination.} The best VLM configurations (progress-description and observe-only) each achieve 67\% forward accuracy (12/18), while CLIP-direct with $\alpha=0.7$ reaches 72\% (13/18). More importantly, CLIP achieves perfect jump detection (6/6) and is $400\times$ faster. The VLM-caption pipeline introduces noise at two stages: first, VLM-generated captions frequently hallucinate task progress (e.g., describing a peg as ``partially inserted'' when it lies on the table at 0\% completion); second, the additional sentence-embedding step (MiniLM) maps semantically distinct captions to similar embedding vectors, further diluting the discrimination signal.

\textbf{The baseline penalty term is theoretically motivated and practically beneficial.} The term $-(1-\alpha)\cdot\mathrm{sim}(f_I(s), f_I(s_0))$ penalizes states that remain visually similar to the initial frame, amplifying the reward gradient in the forward direction. This introduces an expected asymmetry where backward transitions receive less reward, which is acceptable for RL training since episodes always progress forward from the initial state.

\textbf{Completion-sense quality correlates with visual saliency.} Tasks where the goal state is visually distinctive from the initial state (large object displacement, prominent structural changes) produce strong, monotonic potential trajectories. Fine-grained manipulation tasks, where relevant objects are small or partially occluded, yield noisier signals---motivating the use of higher-resolution vision encoders (e.g., ViT-B/16) for such domains.

\textbf{VLM prompting strategy has a dramatic effect.} The evidence-gating prompt achieves 0/18 accuracy because it provides a safe fallback phrase (``No actions completed''), which the VLM defaults to for every frame regardless of actual completion. Progress-description and observe-only prompts achieve identical FTA (12/18), though observe-only achieves perfect jump detection (6/6) by describing scene state more faithfully without goal-echoing bias.

\textbf{CLIP is deterministic and orders of magnitude faster.} CLIP processes a frame in $\sim$5\,ms on GPU, while VLM captioning requires $\sim$2\,s per frame---a $400\times$ speed difference. This enables dense per-step reward computation during RL training, whereas VLM-based rewards must be sparsely sampled (e.g., every 25--50 steps), further disadvantaging VLM approaches for online use.

Based on these results, we adopt the CLIP-direct formulation (Eq.~\ref{eq:clip-potential}) with $\alpha=0.7$ as the default Reward-Zero potential function for all subsequent RL experiments.


\subsection{Embodied Tasks} \label{sec:exp-robot}
\begin{figure}
    \centering
    \includegraphics[width=0.95\linewidth]{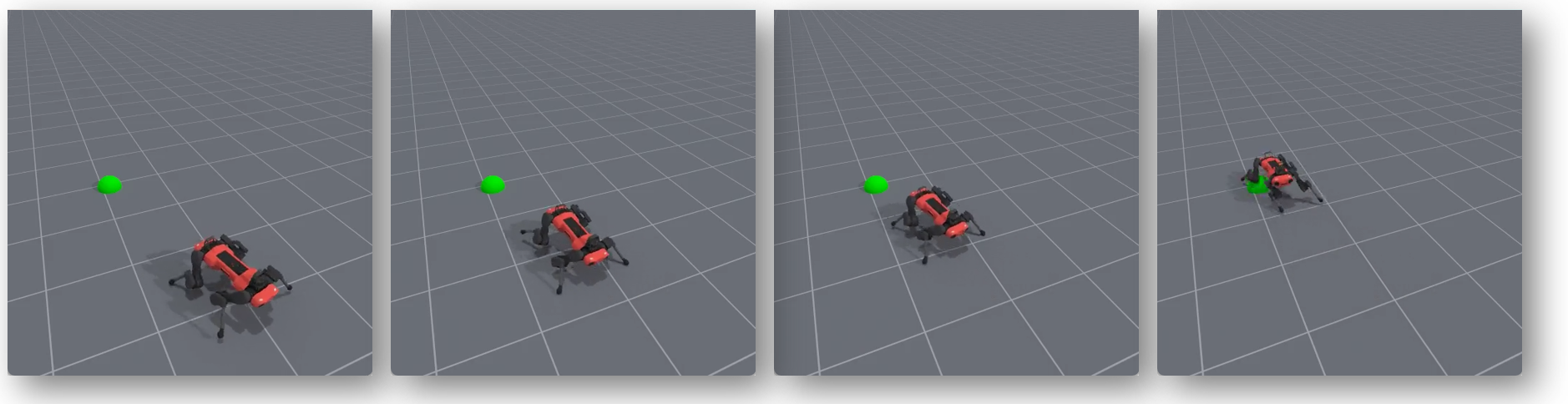}
    \caption{The AnymalC-Reach task involves a quadruped robot learning to navigate to a target location. The task requires the agent to understand spatial relationships and adapt its locomotion strategy accordingly. This environment serves as a challenging testbed for evaluating the effectiveness of our Reward-Zero in guiding learning through language-driven implicit rewards.}
    \label{fig:anymal-reach-task}
\end{figure}

To further deeply evaluate the effectiveness of our proposed Reward-Zero, we conduct experiments on a suite of robotic tasks that require varying levels of complexity and dynamic interactions. 
These tasks are designed to test the generalization capabilities of our method across different scenarios and to compare its performance against standard RL algorithms and common reward-shaping baselines.
One example task is the quadruped robot ``AnymalC-Reach'' task shown in Fig.~\ref{fig:anymal-reach-task}, where the robot must learn to navigate to a target location.

\begin{figure}[h]
    \centering
    \begin{subfigure}[b]{0.32\linewidth}
        \centering
        \includegraphics[width=1\linewidth,trim=0cm 0.7cm 0cm 0.7cm,clip]{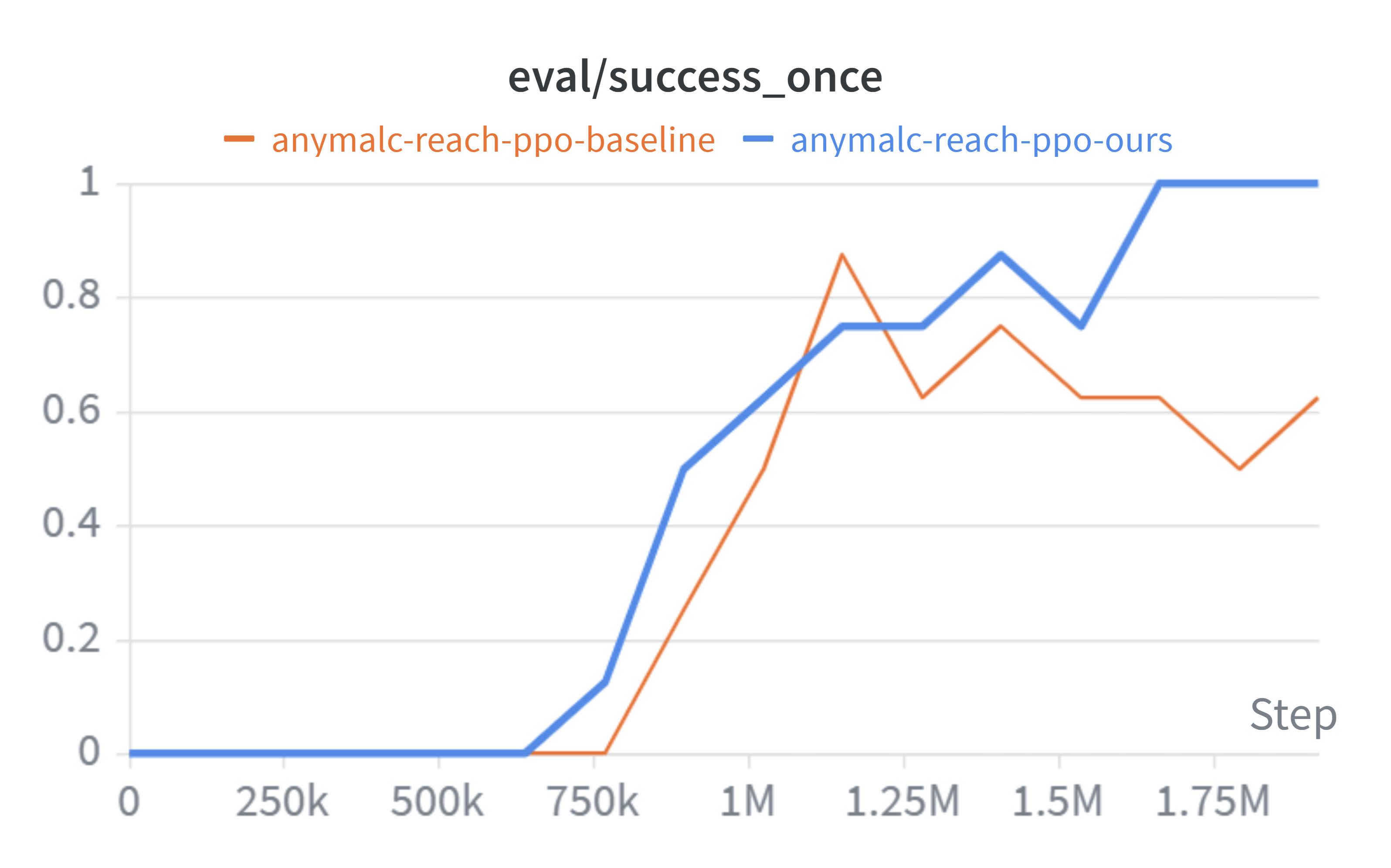}
        \label{fig:anymal-sub1}
    \end{subfigure}
    \hfill
    \begin{subfigure}[b]{0.32\linewidth}
        \centering
        \includegraphics[width=1\linewidth,trim=0cm 0.7cm 0cm 0.7cm,clip]
        {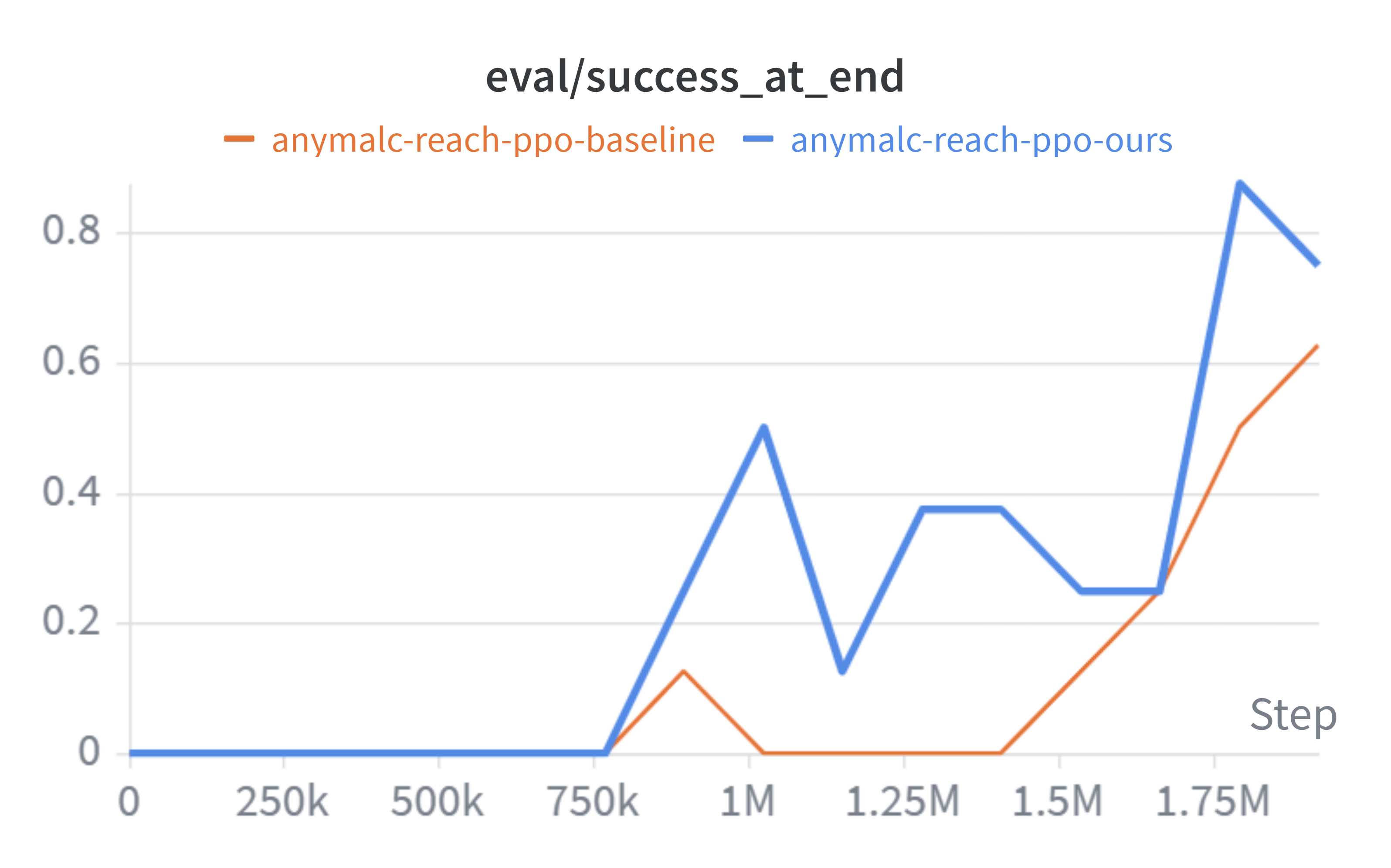}
        \label{fig:anymal-sub2}
    \end{subfigure}
    \hfill
    \begin{subfigure}[b]{0.32\linewidth}
        \centering
        \includegraphics[width=1\linewidth,trim=0cm 0.7cm 0cm 0.7cm,clip]
        {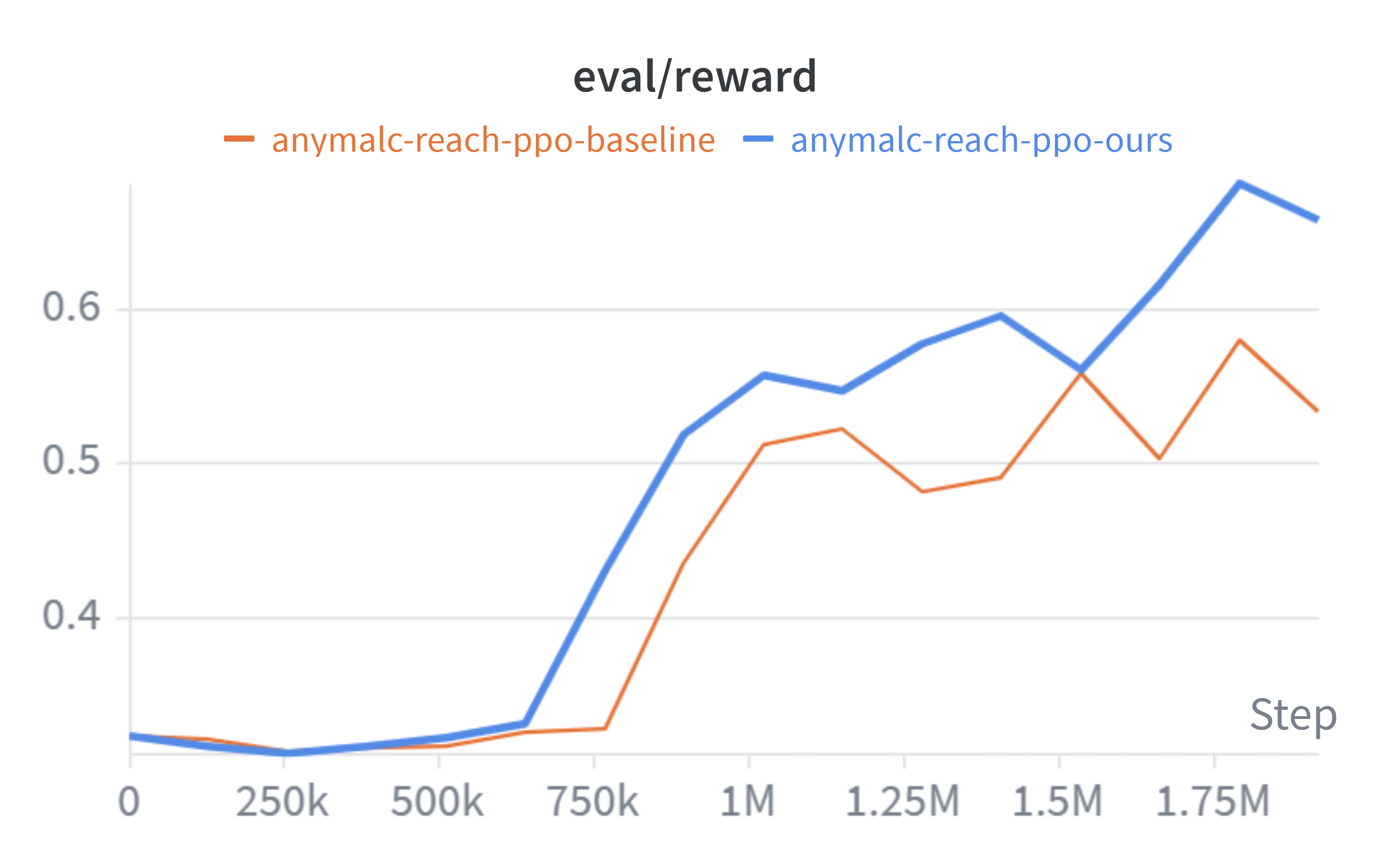}
        \label{fig:anymal-sub3}
    \end{subfigure}
    \vspace{-1.8em}
    \caption{Performance comparison of the ANYmal-C Reach-PPO baseline versus our approach. Solid lines represent the mean values over $2\text{M}$ training steps. Our method significantly outperforms the baseline in both task success rates and cumulative reward.}
    \label{fig:anymal-reach-results}
\end{figure}

We compare the performance of Reward-Zero against traditional RL methods (PPO) and reward-shaping baselines that rely on hand-crafted distance metrics or task-specific features.
Our evaluation metrics include convergence speed, sample efficiency, and final success rate.
The results in Fig.~\ref{fig:anymal-reach-results} demonstrate that Reward-Zero significantly accelerates learning and improves generalization across tasks compared to traditional approaches, highlighting the advantages of our language-driven implicit reward mechanism in guiding agents toward task completion in a more semantically meaningful way.

\begin{figure}[h]
    \centering
    \begin{subfigure}[b]{0.32\linewidth}
        \centering
        \includegraphics[width=1\linewidth,trim=0cm 0.7cm 0cm 0.7cm,clip]{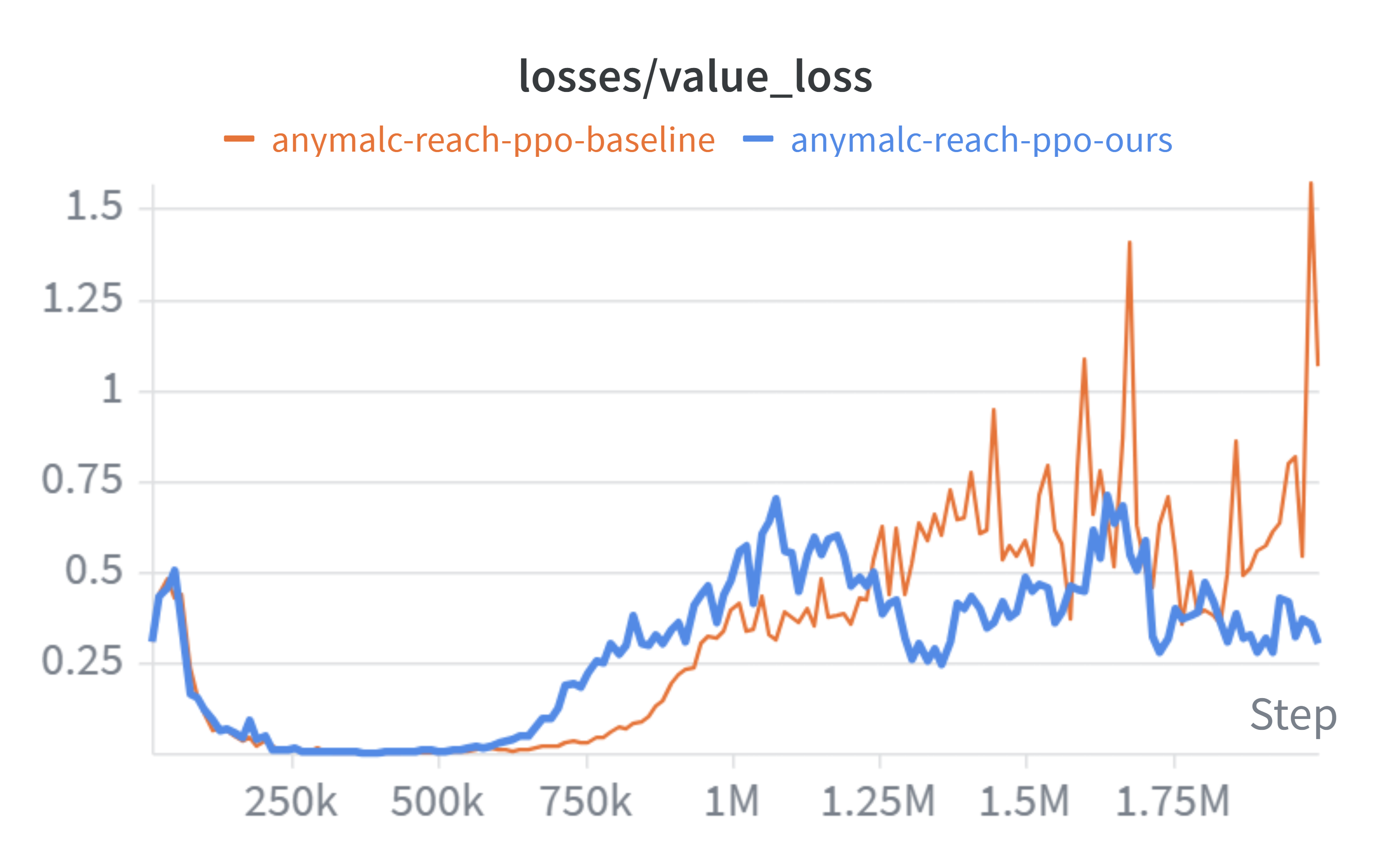}
        \label{fig:dyn-sub1}
    \end{subfigure}
    \hfill
    \begin{subfigure}[b]{0.32\linewidth}
        \centering
        \includegraphics[width=1\linewidth,trim=0cm 0.7cm 0cm 0.7cm,clip]{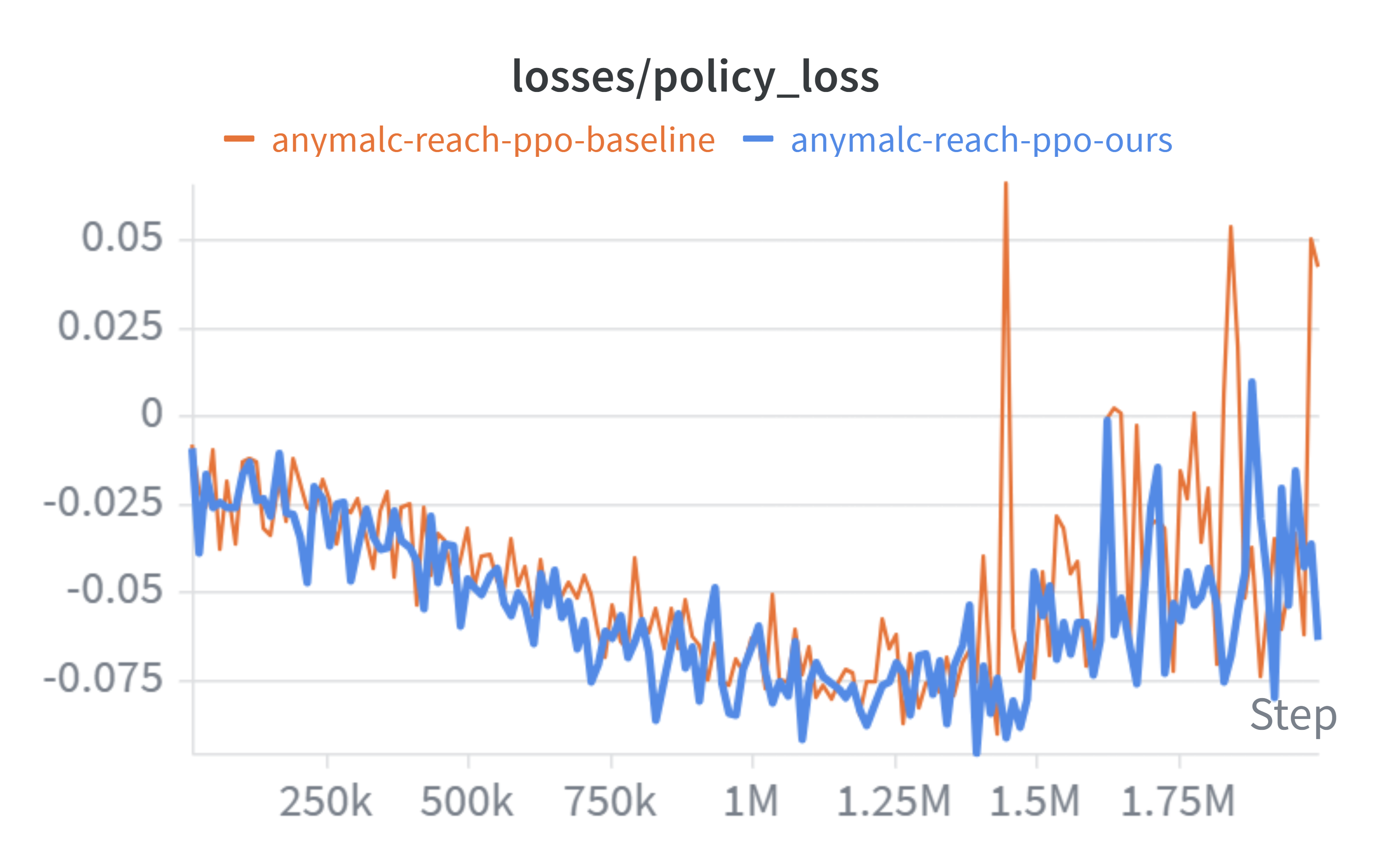}
        \label{fig:dyn-sub2}
    \end{subfigure}
    \hfill
    \begin{subfigure}[b]{0.32\linewidth}
        \centering
        \includegraphics[width=1\linewidth,trim=0cm 0.7cm 0cm 0.7cm,clip]{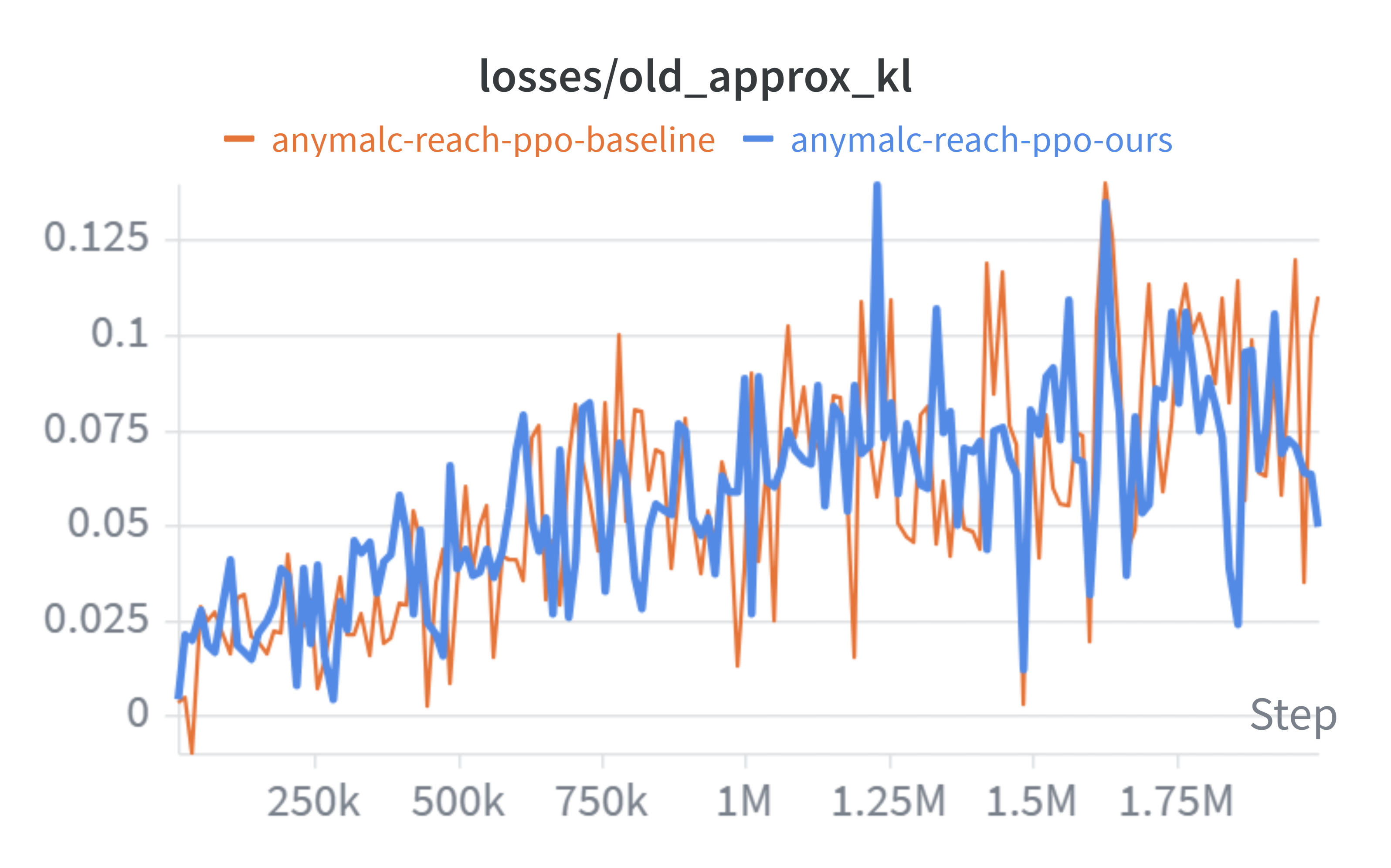}
        \label{fig:dyn-sub3}
    \end{subfigure}
    \\
    \centering
    \begin{subfigure}[b]{0.32\linewidth}
        \centering
        \includegraphics[width=1\linewidth,trim=0cm 0.7cm 0cm 0.7cm,clip]{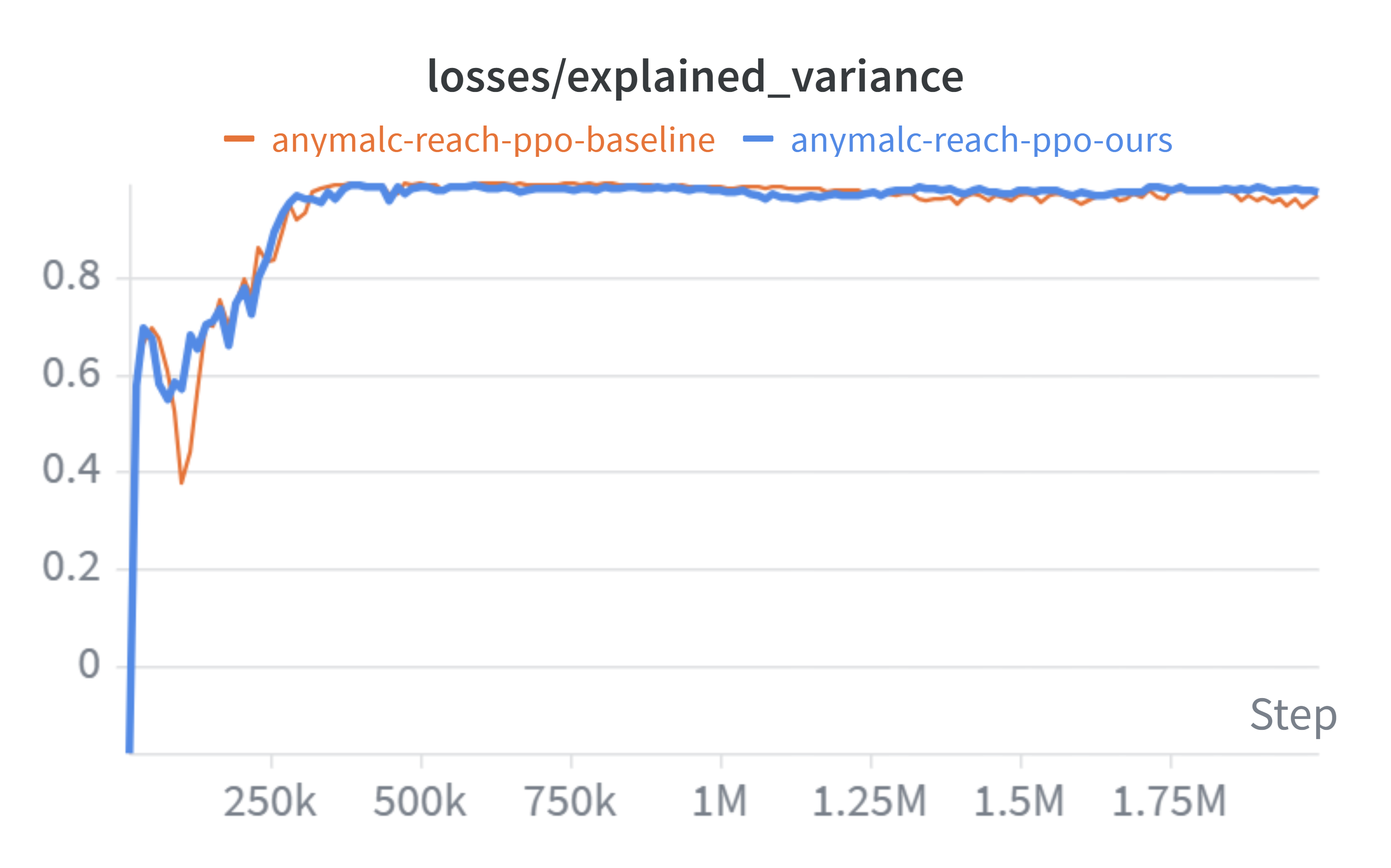}
        \label{fig:dyn-sub4}
    \end{subfigure}
    \hfill
    \begin{subfigure}[b]{0.32\linewidth}
        \centering
        \includegraphics[width=1\linewidth,trim=0cm 0.7cm 0cm 0.7cm,clip]
        {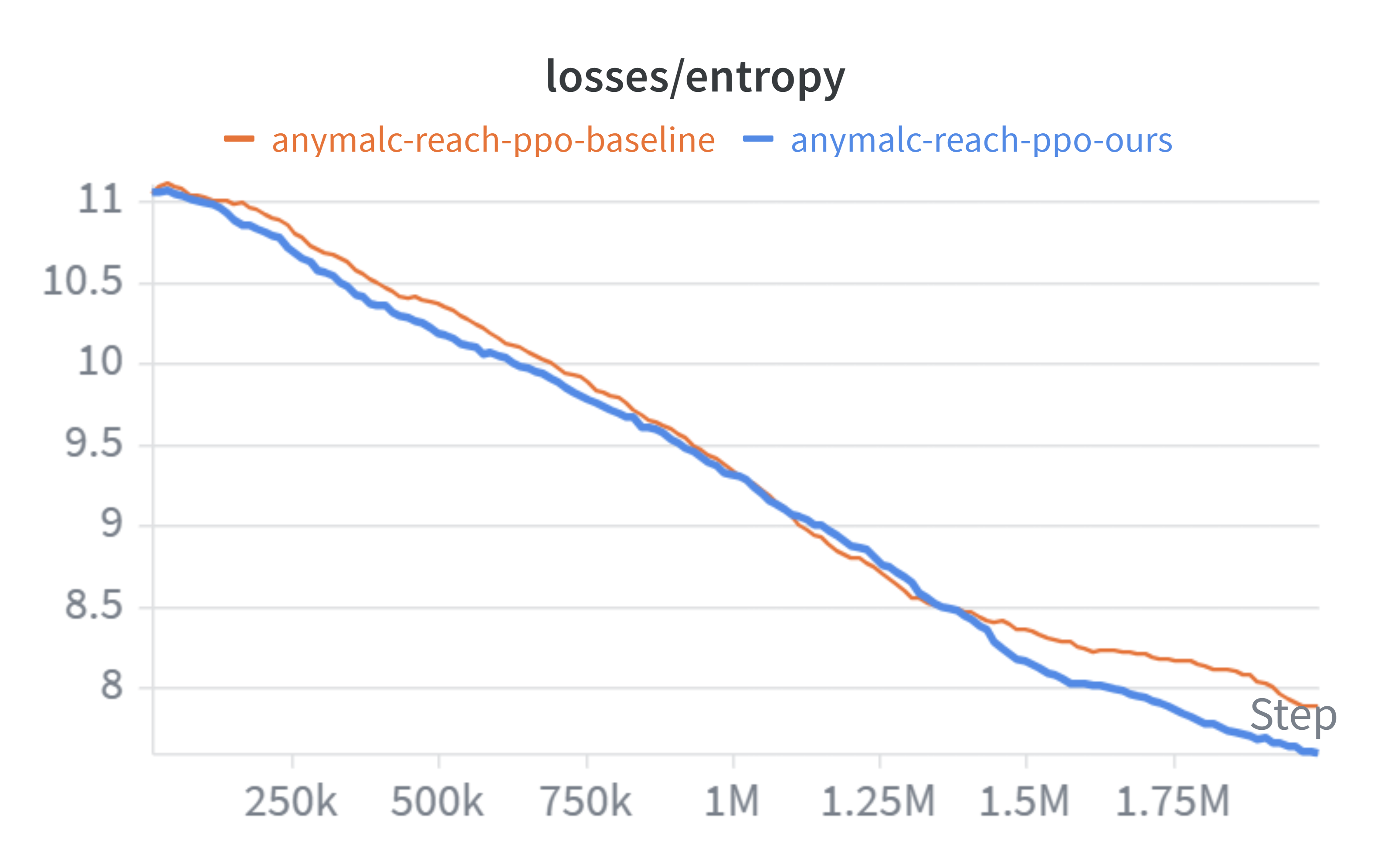}
        \label{fig:dyn-sub5}
    \end{subfigure}
    \hfill
    \begin{subfigure}[b]{0.32\linewidth}
        \centering
        \includegraphics[width=1\linewidth,trim=0cm 0.7cm 0cm 0.7cm,clip]
        {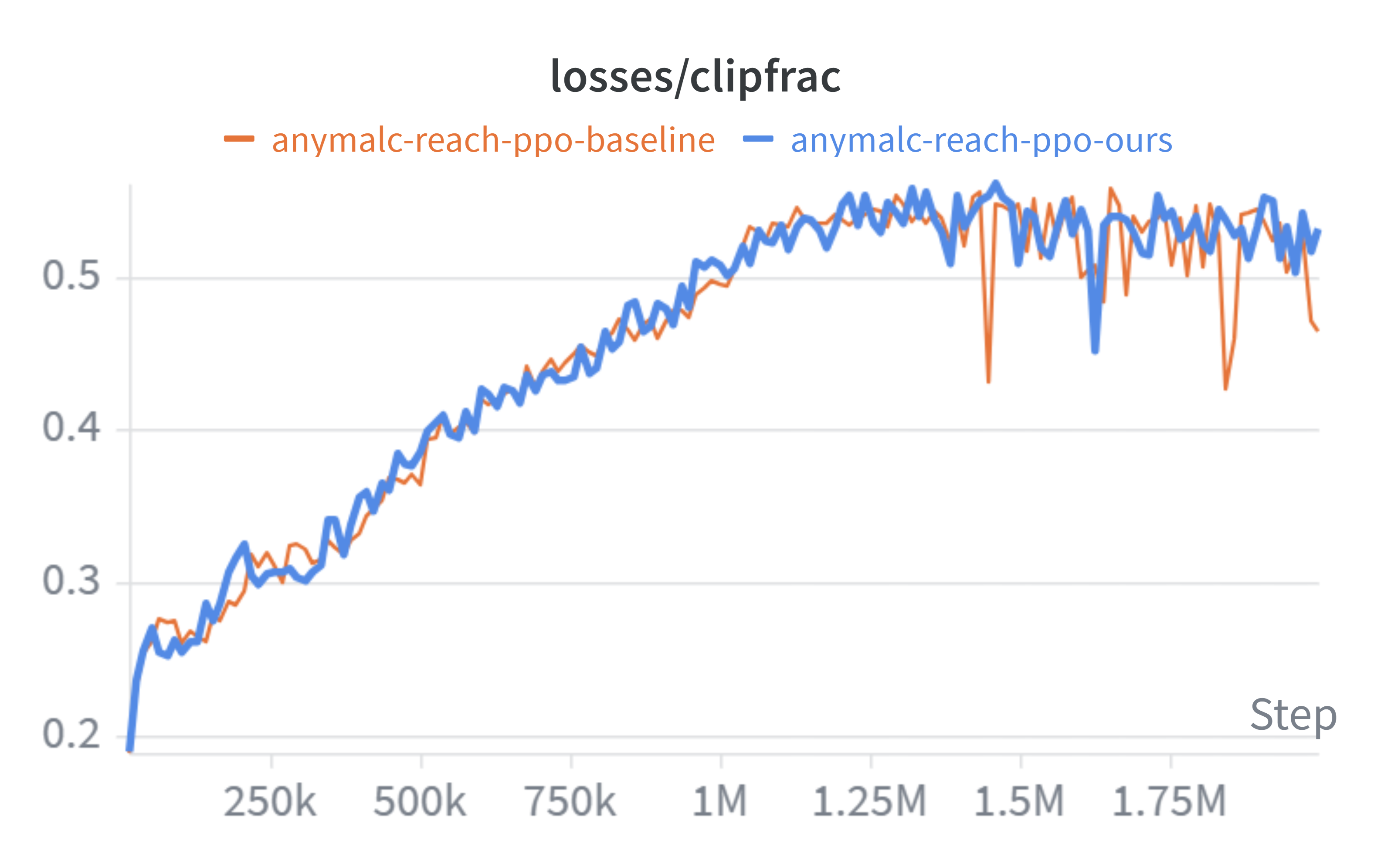}
        \label{fig:dyn-sub6}
    \end{subfigure}
    \vspace{-1.8em}
    \caption{Training dynamics comparison between the PPO baseline (orange) and Ours (blue). The Reward-Zero variant exhibits substantially smoother value loss, more stable policy updates, and more consistent KL and clipping behavior, indicating improved critic reliability and better‑controlled policy optimization. Reward-Zero also shows faster entropy reduction and more stable explained variance, reflecting higher learning efficiency and more reliable advantage estimation throughout training.}
    \label{fig:anymal-reach-results-learning-curves}
\end{figure}

Moreover, we analyze the learning curves in Fig.~\ref{fig:anymal-reach-results-learning-curves} demonstrating that ours yields substantially more stable and reliable optimization dynamics compared to the PPO baseline. 
The value loss of the baseline exhibits pronounced oscillations, particularly in later stages, indicating unstable critic fitting, whereas Reward-Zero maintains a consistently smooth trajectory, reflecting a more accurate and stable value function. Similarly, the policy‑related metrics, including policy loss, KL divergence, and clipping fraction,  show that Reward-Zero performs updates with markedly fewer spikes and reduced variance, 
suggesting more controlled policy improvement and a lower risk of catastrophic updates. In terms of learning efficiency, Reward-Zero achieves faster entropy reduction and maintains higher, more stable explained variance, implying quicker convergence toward effective policies and more reliable advantage estimation. 
Collectively, these results indicate that Reward-Zero not only stabilizes PPO training but also improves sample efficiency and robustness, ultimately enabling more consistent and higher‑quality policy learning.

\subsection{Ablation Study}
\begin{figure}[h]
    \centering
    \begin{subfigure}[b]{0.32\linewidth}
        \centering
        \includegraphics[width=1\linewidth,trim=0cm 0.7cm 0cm 0.7cm,clip]{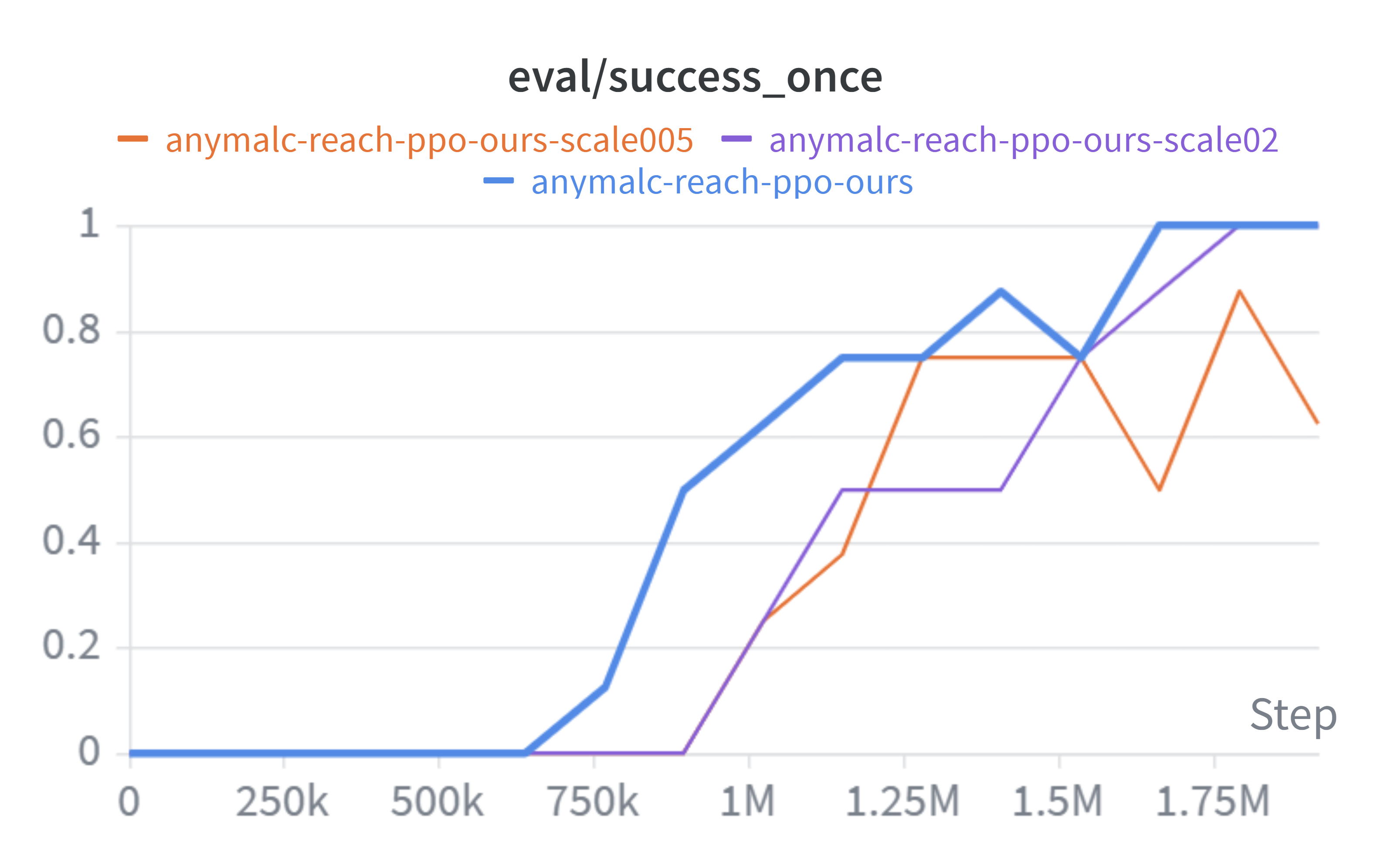}
        \label{fig:abl-scale-sub1}
    \end{subfigure}
    \hfill
    \begin{subfigure}[b]{0.32\linewidth}
        \centering
        \includegraphics[width=1\linewidth,trim=0cm 0.7cm 0cm 0.7cm,clip]
        {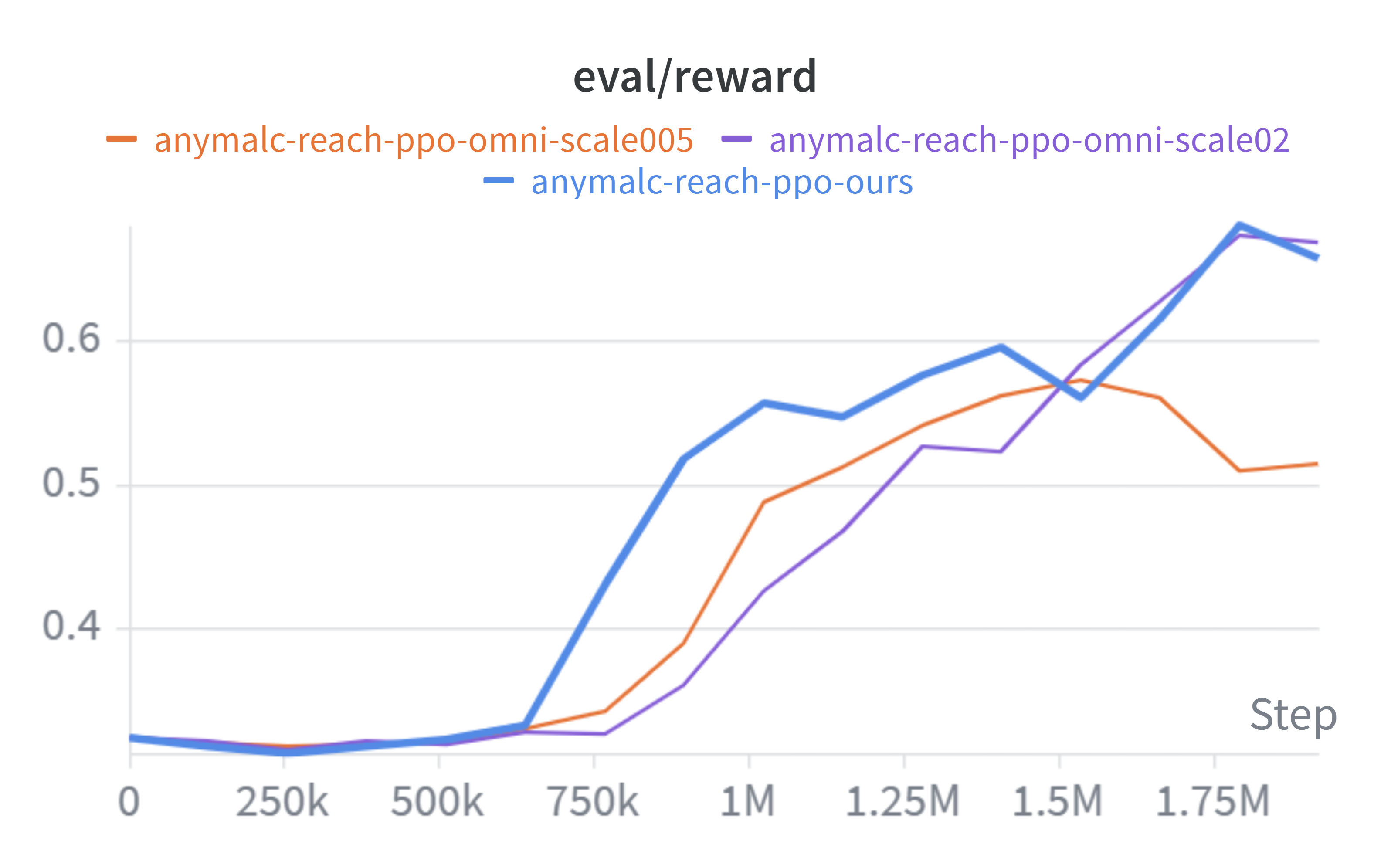}
        \label{fig:abl-scale-sub2}
    \end{subfigure}
    \hfill
    \begin{subfigure}[b]{0.32\linewidth}
        \centering
        \includegraphics[width=1\linewidth,trim=0cm 0.7cm 0cm 0.7cm,clip]
        {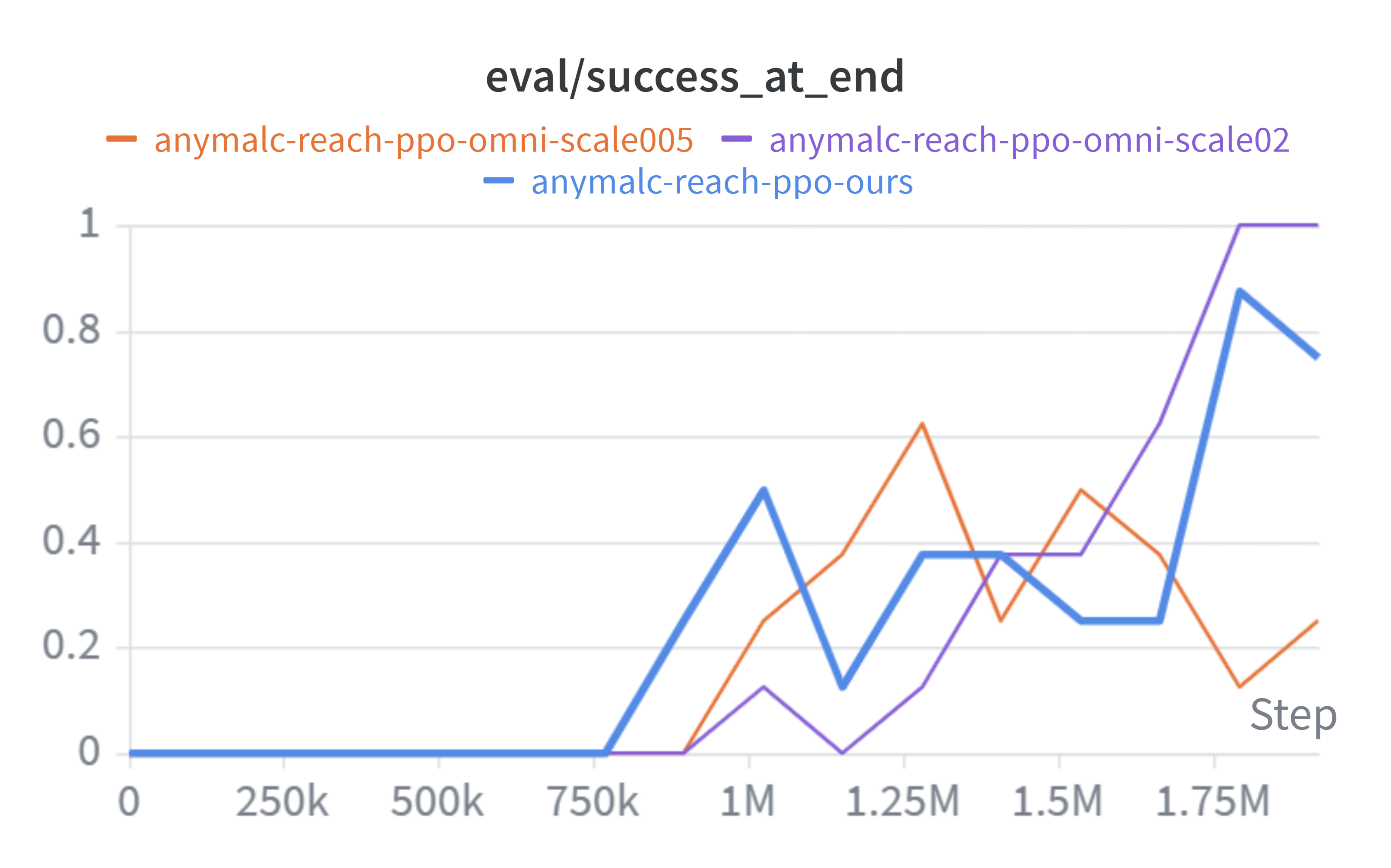}
        \label{fig:abl-scale-sub3}
    \end{subfigure}
    \vspace{-1.8em}
    \caption{Ablation study on the scale parameter for Reward-Zero policy learning.}
    \label{fig:ablation_scale}
\end{figure}
We study the effect of the scale parameter $\beta$ in the Reward-Zero formulation (Eq.~\ref{eq:clip-potential}) on policy learning performance. In Fig.~\ref{fig:ablation_scale}, we compare three configurations: $\beta=0.05$ (low completion bonus, orange), $\beta=0.1$ (default as ours, blue), and $\beta=0.2$ (high completion bonus, purple). 
The results across \texttt{success‑once}, \texttt{success‑at‑end}, and the \texttt{reward}. While reducing or increasing the scale parameter alters learning dynamics, the blue curve (our default) consistently achieves the most reliable and highest overall performance. It reaches perfect \texttt{success‑once} early, maintains strong \texttt{success‑at‑end} behavior, and yields the highest evaluation reward, indicating that the chosen scale value provides the best balance between exploration magnitude and policy update stability. These results highlight that proper scale calibration is crucial for maximizing the effectiveness of completion-sense rewards and ensuring robust task completion.
\begin{figure}[h]
    \centering
    \begin{subfigure}[b]{0.32\linewidth}
        \centering
        \includegraphics[width=1\linewidth,trim=0cm 0.7cm 0cm 0.7cm,clip]{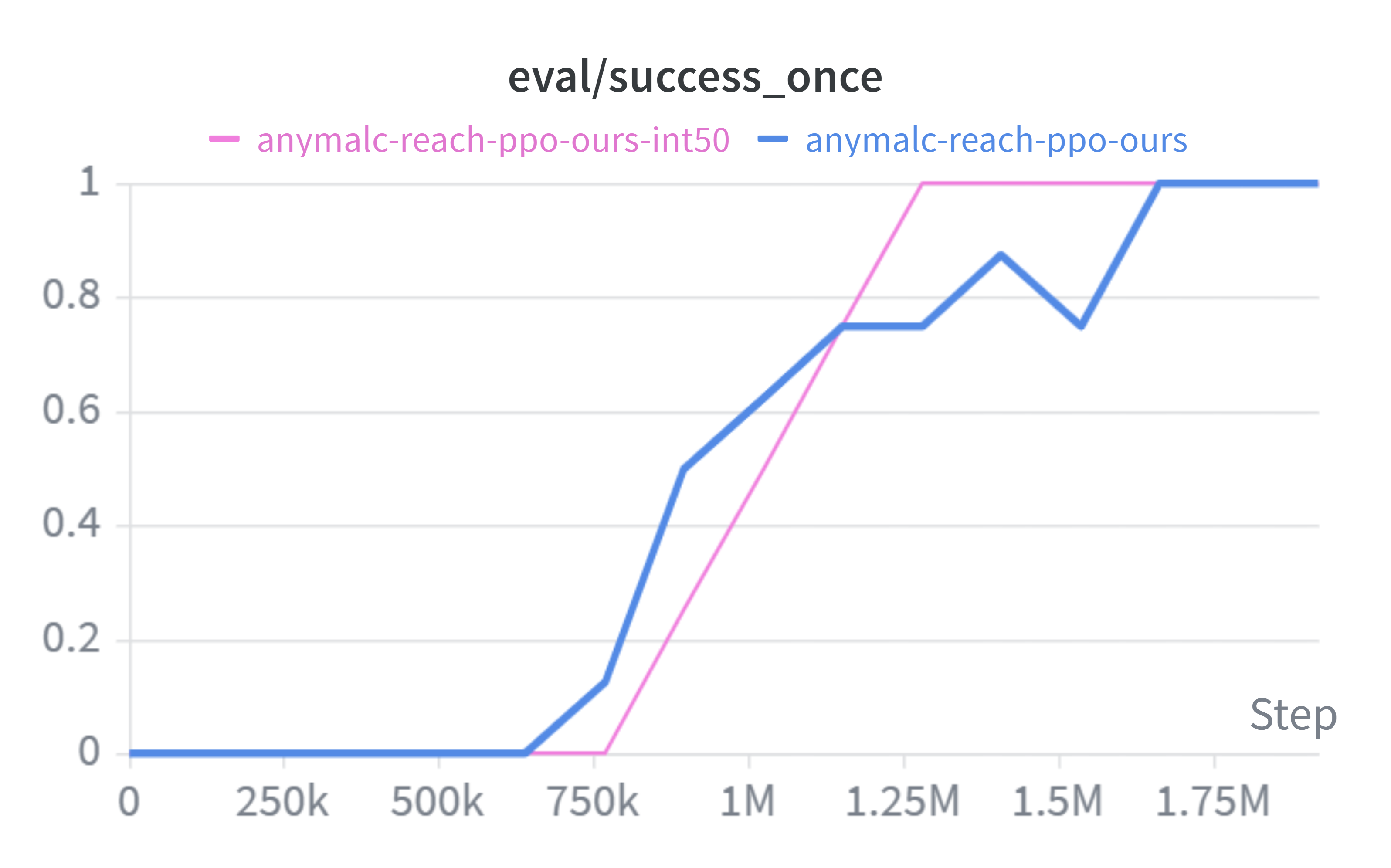}
        \label{fig:abl-int-sub1}
    \end{subfigure}
    \hfill
    \begin{subfigure}[b]{0.32\linewidth}
        \centering
        \includegraphics[width=1\linewidth,trim=0cm 0.7cm 0cm 0.7cm,clip]
        {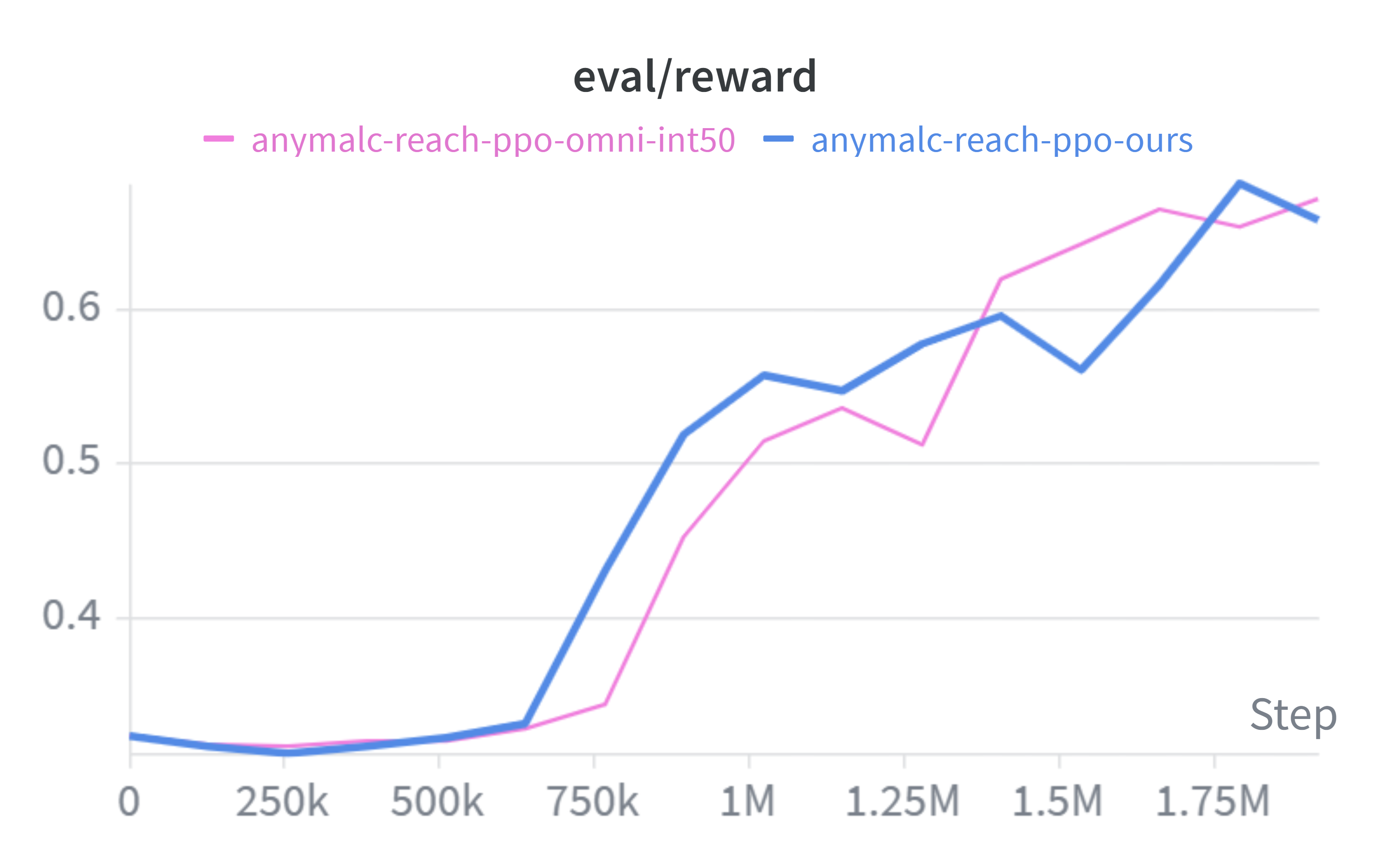}
        \label{fig:abl-int-sub2}
    \end{subfigure}
    \hfill
    \begin{subfigure}[b]{0.32\linewidth}
        \centering
        \includegraphics[width=1\linewidth,trim=0cm 0.7cm 0cm 0.7cm,clip]
        {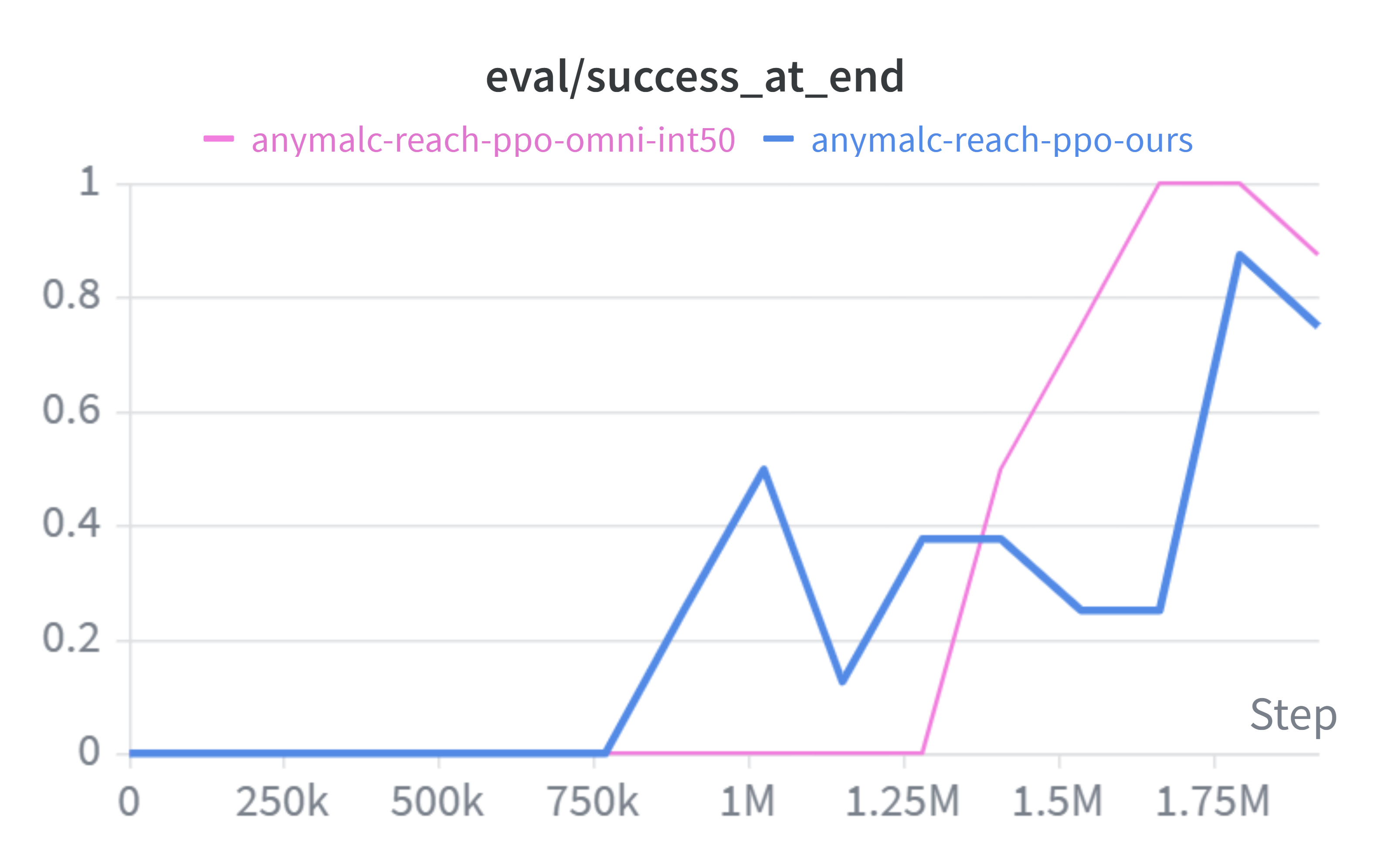}
        \label{fig:abl-int-sub3}
    \end{subfigure}
    \vspace{-1.8em}
    \caption{Ablation on Reward‑Zero Invocation Frequency.}
    \label{fig:ablation_fre}
\end{figure}

We further examine how frequently Reward‑Zero should be invoked during training by comparing our default configuration ($interval=25$) with a less frequent variant ($interval=50$). As shown in Fig.~\ref{fig:ablation_fre}, the results reveal a clear trade‑off between reward‑signal density and policy stability, and they highlight why the default setting provides the most balanced and reliable performance. The evaluation reward curve further clarifies this distinction. Although both variants improve over time, the default interval ultimately achieves slightly higher reward, reflecting more stable policy refinement and better alignment between intermediate progress signals and final task success.

\section{Conclusions and Future Work}
In this paper, we introduced Reward-Zero, a novel reward mechanism for computing dense, semantically-grounded rewards. By leveraging the semantic similarity between scene descriptions and goal specifications, our method provides a powerful potential function that guides learning without requiring task-specific engineering. We demonstrated the effectiveness of Reward-Zero through a mini benchmark evaluating completion-sense discrimination and through experiments on complex robotic tasks, showing improvements in sample efficiency and convergence speed compared to traditional RL approaches.
These results underscore the promise of language‑driven reward shaping for sparse‑reward settings, enabling more generalizable and interpretable reward functions. Future directions include fully language‑embedding–based reward models and success criteria, and deployment in real‑world robotic systems.


\appendix





\bibliography{main}
\bibliographystyle{rlj}

\beginSupplementaryMaterials

\section{Mini benchmark for evaluation of completion sense during task execution} \label{app:bench-frames}

\begin{figure}[h]
\centering
\includegraphics[width=\linewidth]{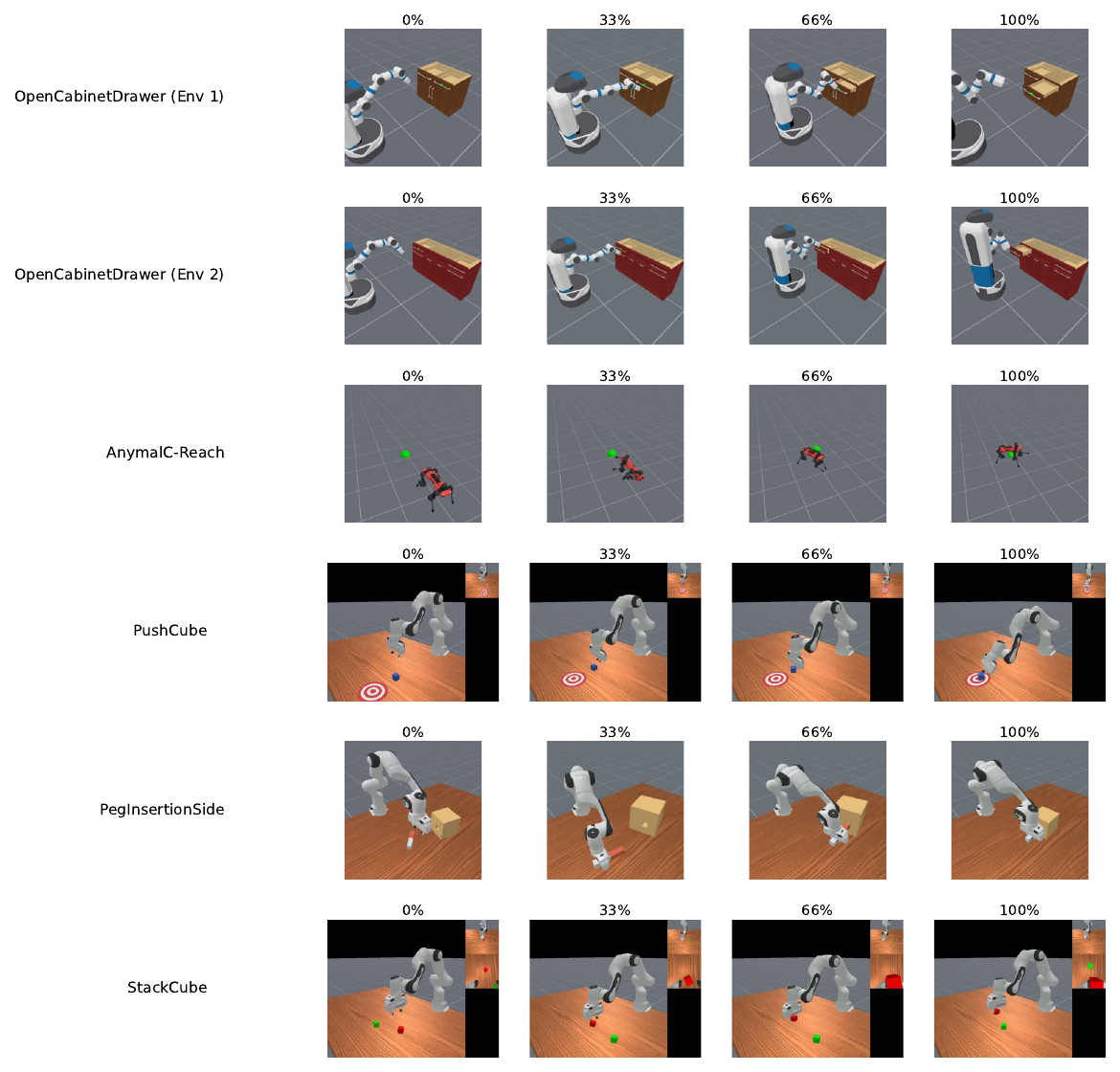}
\caption{Keyframes from the six benchmark episodes (five task types) at four completion stages (0\%, 33\%, 66\%, 100\%). Each row shows one episode; columns correspond to the annotated completion percentages. Tasks range from visually salient changes (top: drawer opening, robot locomotion) to fine-grained manipulation where task-relevant objects are small relative to the scene (bottom: peg insertion, cube stacking).}
\label{fig:bench-frames}
\end{figure}
Figure~\ref{fig:bench-frames} shows representative keyframes for all five tasks at increasing completion stages, and Figure~\ref{fig:bench-curves} plots the corresponding CLIP potential $\Phi(s)$ curves using our best configuration ($\alpha=0.7$). The tasks span a range of visual complexities: OpenCabinetDrawer and AnymalC-Reach involve large, structurally distinctive state changes and produce reliably monotonic potentials, while fine-manipulation tasks (PegInsertionSide, StackCube) involve small objects that occupy only a few pixels at CLIP's 224$\times$224 input resolution, yielding smaller dynamic ranges. Notably, PegInsertionSide's peg ($\sim$2\,cm physical, $\sim$7\,px in CLIP input) fits within a single ViT-B/32 patch, limiting CLIP's ability to track fine-grained insertion progress.

\end{document}